\documentclass[10pt,journal,compsoc,breaklinks=true,colorlinks=true,citecolor=green,urlcolor=magenta,linkcolor=blue,bookmarks=false]{IEEEtran}
\usepackage{booktabs}
\usepackage{graphicx}
\usepackage{multirow}
\usepackage[table]{xcolor}
\usepackage{threeparttable}
\usepackage{todonotes}
\usepackage{array}
\usepackage{amsmath}
\usepackage{amssymb}
\usepackage{float}

\usepackage{booktabs}
\usepackage{tabularx}
\usepackage{array}
\usepackage{makecell}
\usepackage[table]{xcolor} % 若需要斑马线/行底色

\newcolumntype{L}[1]{>{\raggedright\arraybackslash}p{#1}} % 左对齐定宽列
\newcolumntype{C}[1]{>{\centering\arraybackslash}p{#1}}   % 居中定宽列
\newcolumntype{Y}{>{\raggedright\arraybackslash}X}        % 左对齐可伸缩列（tabularx）

\renewcommand\theadfont{\bfseries} % 表头加粗

\usepackage{booktabs}
\usepackage{tabularx}
\usepackage{array}
\usepackage{makecell}
\newcolumntype{L}[1]{>{\raggedright\arraybackslash}p{#1}}
\newcolumntype{Y}{>{\raggedright\arraybackslash}X}

\ifCLASSOPTIONcompsoc
  \usepackage[nocompress]{cite}
\else
  \usepackage{cite}
\fi
\ifCLASSINFOpdf
\else
\fi

\usepackage{booktabs}
\usepackage{graphicx}
\usepackage{multirow} 
\usepackage{microtype}
\usepackage{tikz}
\usepackage[linesnumbered,lined,boxed,commentsnumbered,ruled]{algorithm2e}
\usepackage[breaklinks=true,colorlinks,citecolor=blue,urlcolor=blue,linkcolor=blue,bookmarks=false]{hyperref}
\usepackage[table]{xcolor}
\definecolor{lightgray}{gray}{0.92}
\usepackage{multirow} 
\usepackage{color} 

\usepackage{ragged2e}  % used for align the IEEE abstract

\usepackage{booktabs}
\usepackage{makecell}

\usepackage[capitalize]{cleveref}
\crefname{section}{Sec.}{Secs.}
\Crefname{section}{Section}{Sections}
\Crefname{table}{Table}{Tables}
\crefname{table}{Tab.}{Tabs.}
\crefname{algorithm}{Algo.}{Algos.}

\usepackage{xspace}
\makeatletter
\DeclareRobustCommand\onedot{\futurelet\@let@token\@onedot}
\def\@onedot{\ifx\@let@token.\else.\null\fi\xspace}

\usepackage{forest}
\usepackage{tikz}

\definecolor{darkpastelgreen}{rgb}{0.01, 0.75, 0.24}
\definecolor{lightcoral}{rgb}{0.94, 0.5, 0.5}
\definecolor{myblue}{HTML}{BCDBF6}
\definecolor{mygreen}{rgb}{0.678, 0.875, 0.690}
\definecolor{myred}{rgb}{0.949, 0.792, 0.780}
\definecolor{myyellow}{rgb}{0.976, 0.953, 0.600}
\definecolor{mycream}{rgb}{1.000, 0.957, 0.765}

\usepackage{xcolor}     % 用于定义颜色
\usepackage{graphicx}   % 插图用（可选）
\usepackage{hyperref}   % 超链接支持（必须）

\hypersetup{
    colorlinks = true,        % 启用彩色文本而不是边框
    citecolor  = blue,        % 文献引用的颜色
    linkcolor  = red,         % \ref, \autoref 等链接的颜色（即图表、章节引用）
    urlcolor   = green,       % URL 链接颜色（可选）
    filecolor  = magenta      % 文件链接颜色（可选）
}

\usepackage{enumitem}

\begin{document}
\title{Cognitive Pivot Points and Visual Anchoring: Unveiling and Rectifying Hallucinations in Multimodal Reasoning Models}

\author{
Zhe~Qian,
Yanbiao~Ma{\textsuperscript\dag},
Zhuohan~Ouyang,
Zhonghua~Wang,
Zhongxing~Xu,
Fei~Luo,
%Chenyue~Zhou,
%Yanxi~Lu,
%Mingxuan~Wang,
Xinyu~Liu,
Zongyuan Ge,~\IEEEmembership{Senior member,~IEEE},
Yike~Guo,~\IEEEmembership{Fellow,~IEEE},~and~Jungong Han~\IEEEmembership{Senior member,~IEEE}%

\thanks{ Yanbiao Ma is with the Gaoling School of Artificial Intelligence, Renmin University of China. Zhe Qian is with South China Agricultural University. Zhuohan Ouyang is with South China Normal University. Zhonghua Wang, Zhongxing Xu, and Zongyuan Ge are with Monash University. Fei Luo is with Jishou University. Xinyu Liu and Yike Guo are with the Hong Kong University of Science and Technology. Jungong Han is with Tsinghua University.}%
\thanks{\textsuperscript{\dag}Corresponding author: Jungong Han. E-mail: ybma1998@ruc.edu.cn, jungonghan77@gmail.com}%
}


\markboth{{Journal of \LaTeX\ Class Files, September 2026}}%}%
{Shell \MakeLowercase{\textit{et al.}}: Bare Advanced Demo of IEEEtran.cls for IEEE Computer Society Journals}

\IEEEtitleabstractindextext{
\begin{abstract}
%\justifying

Multimodal Large Reasoning Models (MLRMs) have achieved remarkable strides in visual reasoning through test time compute scaling, yet long chain reasoning remains prone to hallucinations.
We identify a concerning phenomenon termed the Reasoning Vision Truth Disconnect (RVTD): hallucinations are strongly correlated with cognitive bifurcation points that often exhibit high entropy states. We attribute this vulnerability to a breakdown in visual semantic anchoring, localized within the network's intermediate layers; specifically, during these high uncertainty transitions, the model fails to query visual evidence, reverting instead to language priors. Consequently, we advocate a shift from solely outcome level supervision to augmenting it with fine grained internal attention guidance. To this end, we propose V-STAR (Visual Structural Training with Attention Reinforcement), a lightweight, holistic training paradigm designed to internalize visually aware reasoning capabilities. Central to our approach is the Hierarchical Visual Attention Reward (HVAR), integrated within the GRPO framework. Upon detecting high entropy states, this mechanism dynamically incentivizes visual attention across critical intermediate layers, thereby anchoring the reasoning process back to the visual input. Furthermore, we introduce the Forced Reflection Mechanism (FRM), a trajectory editing strategy that disrupts cognitive inertia by triggering reflection around high entropy cognitive bifurcation points and encouraging verification of subsequent steps against the visual input, thereby translating external debiasing interventions into an intrinsic capability for hallucination mitigation.

\end{abstract}

}

\maketitle

% is not selected <OR> if conference mode is selected - because compsoc
% conference papers position the abstract like regular (non-compsoc)
\IEEEdisplaynontitleabstractindextext

% \fi
\IEEEpeerreviewmaketitle

\section{Introduction}
\label{1}

\begin{figure}[t]
    \centering
    \hspace*{-2.0mm}\includegraphics[width=1.0\linewidth]{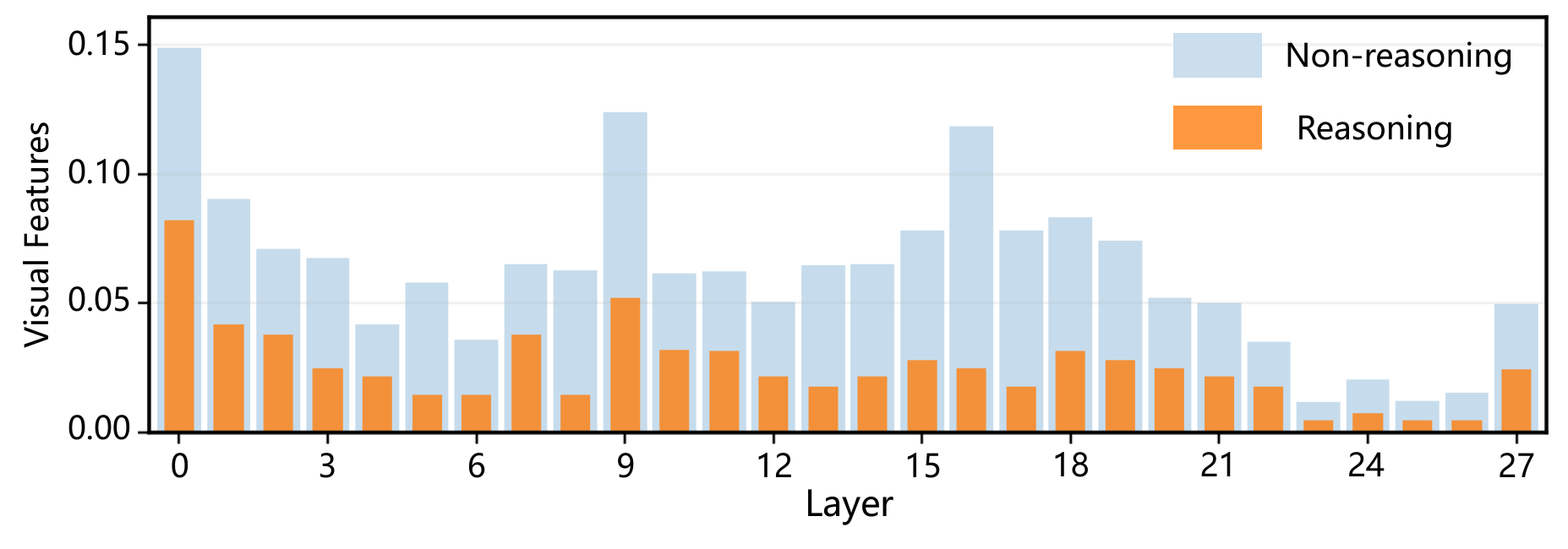}
    \vskip -0.2in
    \caption{\textbf{Reasoning relies less on visual evidence.}
    Across layers, reasoning generations show reduced visual feature signals compared to non-reasoning.}
    \label{fig1}
\vskip -0.2in
\end{figure}

% --- Paragraph 1: Context & The Paradox ---
\IEEEPARstart{T}{he} advent of Large Reasoning Models (LRMs)~\cite{paper2, paper12, paper41, paper34,444} has marked a paradigm shift in artificial intelligence, enabling models to solve complex problems through test time compute scaling~\cite{paper1, paper11} and extensive Chains of Thought (CoT)~\cite{paper3, paper10, paper39}. Recently, this paradigm has been successfully adapted to the multimodal domain~\cite{paper4, paper5, paper13, paper14, paper15, paper16, paper17,tpami1}. Multimodal Large Reasoning Models (MLRMs)~\cite{paper8} achieve a deep synthesis of visual perception and linguistic reasoning by constructing explicit reasoning chains reinforced by verifiable rewards~\cite{paper22, paper75}. However, despite these impressive capabilities, we identify a fundamental vulnerability termed the ``Reasoning Vision Truth Disconnect": compared to non reasoning generation, chain of thought reasoning typically expands the generation trace into longer sequences, making models more susceptible to hallucination~\cite{paper7, paper18, paper20, paper21, paper22,tpami2} by underutilizing visual evidence. Our empirical analysis reveals a counterintuitive trend (Fig.~\ref{fig1}): MLRMs attend to visual evidence markedly less in chain of thought reasoning than in non-reasoning generation, showing a consistently lower level of visual attention throughout the network. This suggests a mode level shift toward internal linguistic inference and away from visual grounding.\cite{tpami3}

% --- Figure 1: Phenomenon ---

% --- 
\begin{figure*}[t]
    \vskip -0.0in
    \centering
    \centerline{\includegraphics[width=1.00\textwidth]{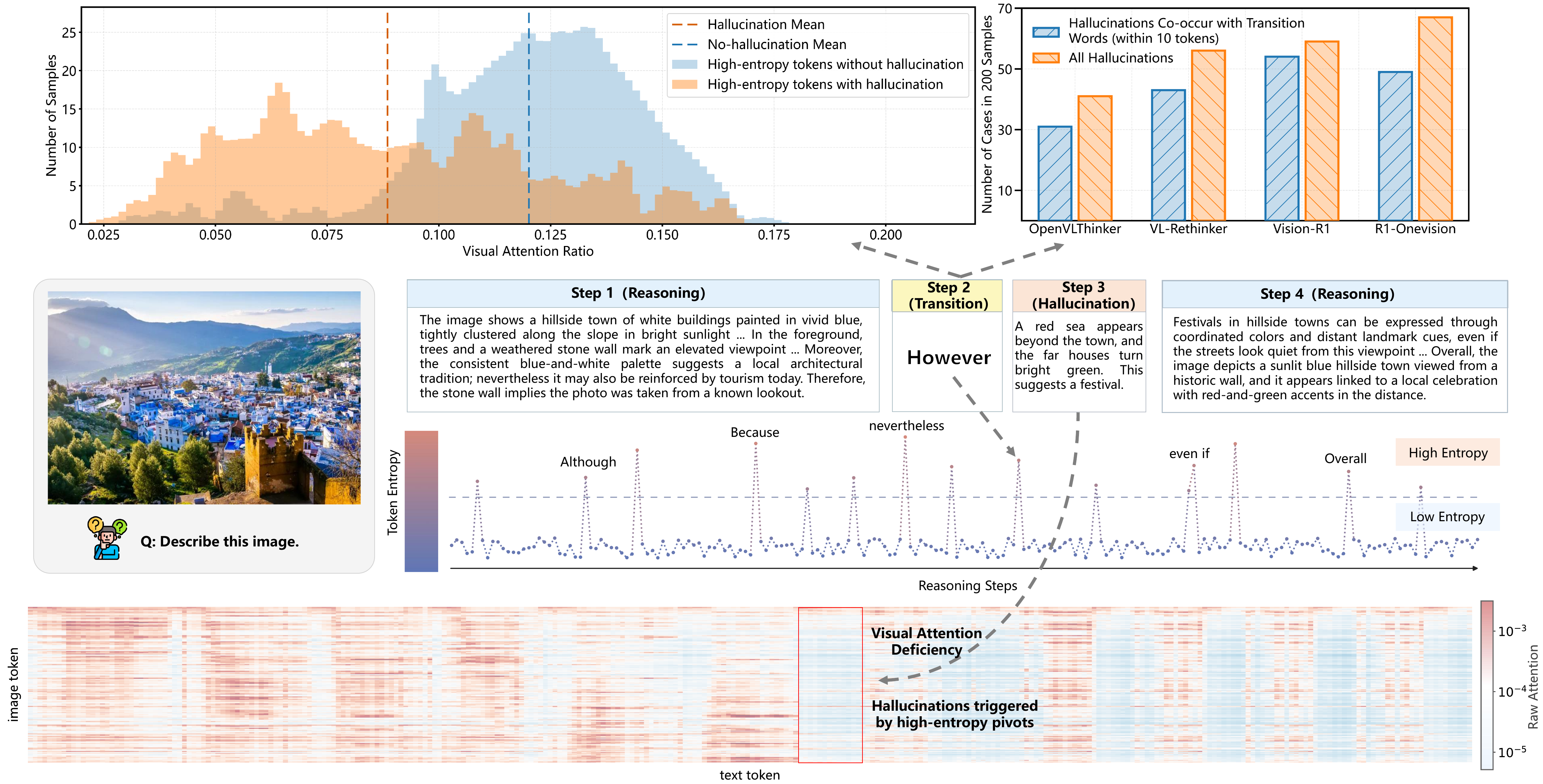}} 
    \vskip -0.1in
    \caption{\textbf{A multi view analysis of hallucination triggers in multimodal reasoning.} The statistics show that hallucinations tend to occur near high entropy transition words such as “However”, and the visual attention ratio is generally lower in these cases. In the example trajectory, “However” coincides with a spike in token entropy and is immediately followed by content that contradicts the image. The token level attention heatmap further indicates a strong suppression of attention to image tokens around this high entropy transition, after which the generation enters a hallucinated span highlighted by the red box, suggesting a temporary loss of visual grounding at the transition. }
    \label{fig2}
    \vskip -0.0in
\end{figure*}
%{figure/intro111111111111111111.pdf}} 

To identify the root cause of this phenomenon, we conducted a deep analysis of the internal mechanisms of MLRMs. As shown in Fig.~\ref{fig2}, transition words (such as however and wait) act as cognitive bifurcation points in the reasoning chain. At this stage, semantic uncertainty sharply increases (manifesting as a high entropy state), and hallucinations frequently occur together with these transition steps. Crucially, we discovered a severe mismatch. Exactly at these bifurcation points, where visual verification is most needed to resolve uncertainty, the model fails to allocate enough visual attention. As demonstrated by the attention heatmap Fig.~\ref{fig2} of text and visual tokens in the figure, the attention from text tokens to image tokens becomes notably weak around the high entropy transition positions (red boxes). Furthermore, hallucinated segments often appear right after these transition points, causing the model to lean more toward language priors to continue its reasoning process.

\begin{figure*}[t]
    \vskip -0.00in
    \centering
    \centerline{\includegraphics[width=2.05\columnwidth]{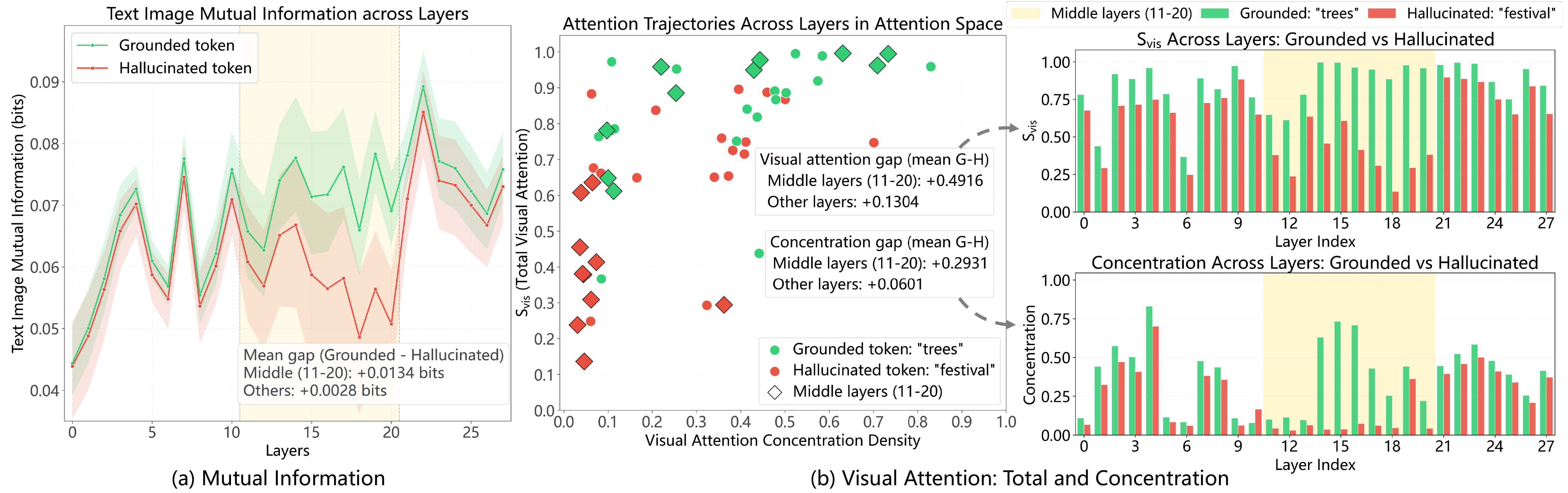}}
    \vskip -0.1in
    \caption{\textbf{Intermediate layer divergence between grounded and hallucinated tokens.}
    (a) \textbf{Text Image Mutual Information:} The gap between grounded and hallucinated tokens peaks in the intermediate layers from 11 to 20, where grounded tokens preserve higher Text Image Mutual Information.
    (b) \textbf{Visual attention total and concentration:} In the same intermediate layer window, grounded tokens exhibit higher total visual attention $S_{\mathrm{vis}}$ and higher visual attention concentration, while hallucinated tokens show reduced $S_{\mathrm{vis}}$ together with a drop in concentration, indicating a redistribution away from visual evidence in intermediate layers.}
    \label{fig3}
    \vskip -0.00in
\end{figure*}

\begin{figure}[htbp]
    \centering
    \vskip -0.055in
    \hspace*{-2.4mm}\includegraphics[width=1.01\linewidth]{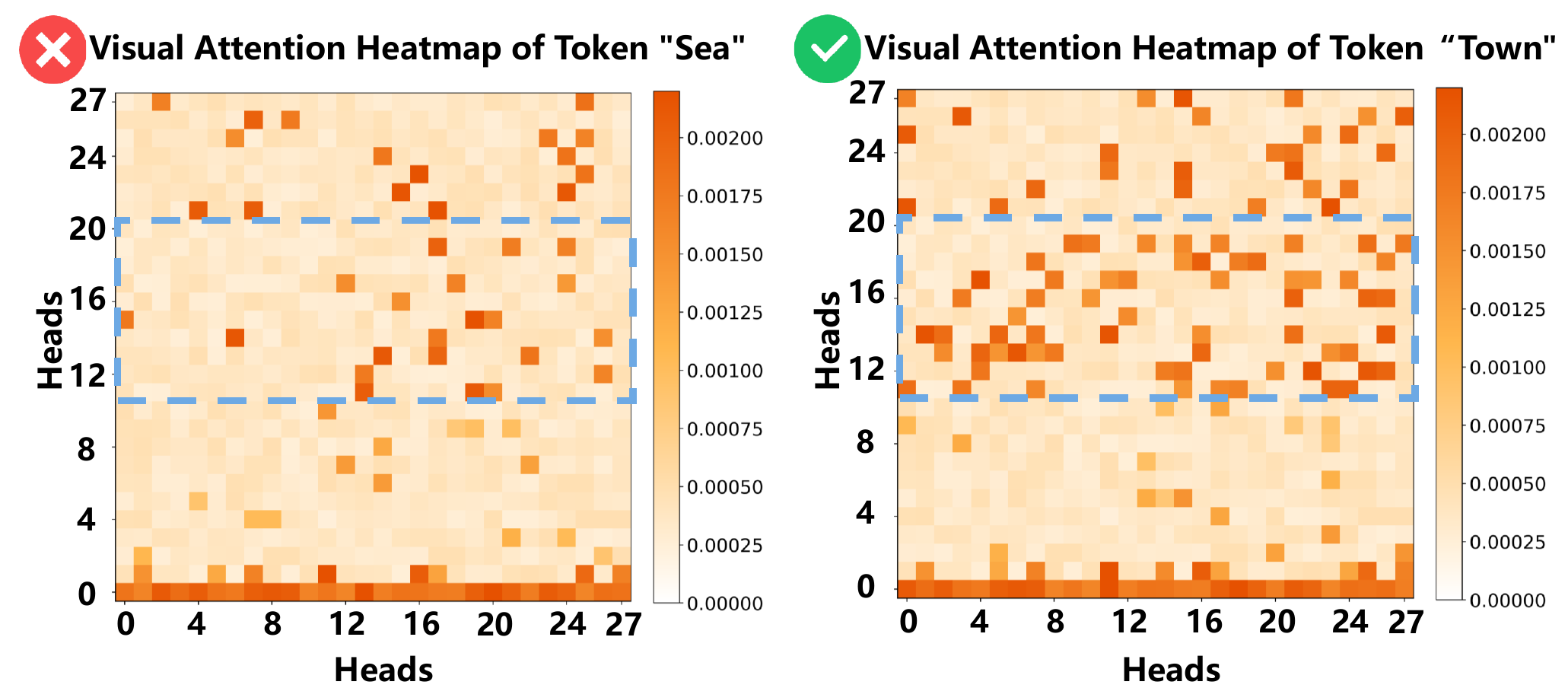}
    \vskip -0.1in
    \caption{\textbf{Head wise visual attention differs for grounded and hallucinated tokens.}
    Visual attention heatmaps compare the hallucinated token “Sea” and the grounded token “Town”. In the highlighted intermediate layer region, the grounded token shows stronger and more coherent attention across heads, while the hallucinated token exhibits weaker and more sporadic activation, indicating reduced visual grounding.}
    \label{fig4}
    \vskip -0.05in
\end{figure}

This macroscopic attention deficit compelled us to examine the model’s microscopic operations: at which stage is visual information lost? Our layer wise analysis (see Finding~2, Sec.~\ref{3.4}) shows that the shallow and deep layers each remain relatively stable, while the true divergence emerges in the intermediate layers. As shown in Fig.~\ref{fig3}(a), the text–image mutual information gap between grounded and hallucinatory tokens increases markedly in the intermediate layers, indicating that the two token types rapidly separate in whether their textual representations remain coupled with visual evidence; for hallucinatory tokens, the strength of cross modal binding clearly decays. 
Furthermore, Fig.~\ref{fig3}(b) provides more direct evidence from attention in the same range: in the intermediate layers, grounded tokens not only allocate more total visual attention, but also exhibit more concentrated attention that points to relevant regions; in contrast, hallucinatory tokens show weaker and more diffuse visual attention, and are more easily influenced by background content or language priors. Finally, the layer by head comparison of visual attention (Fig.~\ref{fig4}) corroborates this localization at a finer granularity: in the intermediate layers, truthful generations display stronger and more consistent high score activations across multiple heads, whereas hallucinatory generations exhibit weaker and more sporadic head level responses. Taken together, these observations allow us to more precisely characterize the model’s microscopic behavior: visual information is lost due to the failure of “Visual Semantic Anchoring” in the intermediate layers, which prevents visual evidence from being stably bound to the target token representations and therefore from being effectively injected into subsequent reasoning.

% --- Figure 3: Heatmaps ---

% --- Paragraph 4: The Solution (V-STAR) ---
To address this disconnectedness, we propose \textbf{V-STAR} (Visual Structural Training with Attention Reinforcement), a holistic training paradigm designed to internalize visually aware reasoning capabilities. First, we integrate a \textbf{Hierarchical Visual Attention Reward (HVAR)} into the GRPO framework. Informed by our discovery that intermediate layers govern anchoring, HVAR employs a dual track strategy: it specifically incentivizes attention in anchoring relevant intermediate layers during high entropy bifurcation points, enforcing precise visual grounding when it is most needed. Second, to counteract the cognitive inertia that prevents models from self correcting, we introduce the \textbf{Forced Reflection Mechanism (FRM)}. This trajectory intervention strategy simulates human deliberation by constructing "Generate, Reflect, and Regenerate" trajectories during RL exploration. By inserting explicit reflection triggers following initial responses, we compel the model to reexamine visual evidence, effectively translating external debiasing interventions into an intrinsic antihallucination capability.

% --- Paragraph 5: Contributions ---
This paper advocates a shift from external decoding interventions to a new paradigm of microscopic guidance and intrinsic debiasing via internal attention allocation. Our primary contributions are summarized as follows:

\begin{itemize}
\item \textbf{Cognitive Pivot Analysis:} We systematically characterize the internal mechanisms of "Cognitive Bifurcation Points," revealing that the fatal mismatch between high uncertainty and low visual attention at transition tokens is the macroscopic driver of hallucination (Fig.~\ref{fig2}), especially at key transitions.

\item \textbf{Layer wise Interpretability:} We pinpoint the decisive role of intermediate layers in "Visual Semantic Anchoring." Our analysis confirms that visual information loss stems from abnormal visual silence in these layers rather than top level failure (Fig.~\ref{fig3}), providing a microscopic empirical basis for fine grained governance.

\item \textbf{Hierarchical Visual Reward:} We propose the Hierarchical Visual Attention Reward (HVAR) within the GRPO framework. By creatively utilizing intermediate layer attention as a reward signal for high entropy points, we achieve precise visual reinforcement during critical decision phases without disrupting reasoning.

\item \textbf{Active Self Correction \& Scalability:} We introduce the Forced Reflection Mechanism (FRM) to internalize deliberative reasoning. Extensive experiments on a newly constructed dataset of 40,000 instructions demonstrate that V-STAR achieves State of the Art performance across multiple multimodal reasoning benchmarks and diverse evaluation settings.
\end{itemize}

\section{Related Work}
\label{2}

This section reviews existing research from the unified perspective of optimizing long-chain multimodal reasoning processes and addressing hallucination challenges.

\textbf{Reasoning with Reinforcement Learning} 
The advent of GPT-o1~\cite{paper1} and DeepSeek-R1~\cite{paper2} has catalyzed a surge of interest in leveraging Reinforcement Learning (RL)~\cite{paper45, paper46} to elicit long Chain of Thought (CoT) generation~\cite{paper3, paper10} and autonomous self-correction~\cite{paper30} in Large Language Models (LLMs)~\cite{paper42}. Beyond these recent reasoning-oriented systems, this line of work is also rooted in classical RL and alignment foundations, including Proximal Policy Optimization (PPO)~\cite{paper33}, RLHF-style instruction following~\cite{paper34}, and GRPO-based reasoning optimization introduced in DeepSeekMath~\cite{paper6}. GPT-o1 exemplified the transformative potential of large-scale RL in inducing slow-thinking capabilities~\cite{paper1}, while DeepSeek-R1 demonstrated that sparse cold-start data, coupled with Group Relative Policy Optimization (GRPO)~\cite{paper6}, can empower models to spontaneously evolve highly intricate reasoning pathways for mathematical and logical tasks~\cite{paper43}. Moreover, studies such as Reflexion~\cite{paper30} and Self-Refine~\cite{paper31} further suggest that verbal reflection and iterative self-feedback can substantially enhance test-time reasoning quality. In parallel, works like SimpleRL-Zoo~\cite{paper55} have probed the feasibility of direct RL from base models, substantiating the latent potential for autonomous evolution sans dense Supervised Fine-Tuning (SFT)~\cite{paper56}. Regarding algorithmic refinement, approaches such as DAPO~\cite{paper57}, Dr GRPO~\cite{paper58}, and DeepScaler~\cite{paper59} address endemic RL challenges like reward hacking~\cite{paper60} and advantage estimation instability~\cite{paper61}. By incorporating robust reward structures and regularization operators~\cite{paper62}, these methods compel models to adhere to rigorous logical closures rather than speculative generation. V-STAR expands this frontier by addressing the coarse-grained nature of intervention inherent in prior frameworks, extending supervisory signals from macroscopic outcome feedback to microscopic reasoning nodes, thereby achieving precise intervention at the focal points of reasoning failure.

\medskip

\textbf{Multimodal Reinforcement Learning} 
Integrating RL into multimodal domains~\cite{paper63} to bolster cross-modal reasoning is a burgeoning research vector~\cite{paper65}~\cite{111}. Prior to RL-centric multimodal reasoning, Multimodal-CoT~\cite{paper5} and more recent step-wise reasoning frameworks such as LLaVA-CoT~\cite{paper4} showed that staged rationale generation and structured multistage reasoning can substantially improve visual problem solving. This direction is also closely related to visual grounding, which emphasizes anchoring linguistic reasoning to localized visual evidence rather than relying solely on coarse global semantics~\cite{tpami4,tpami6,777}. Drawing inspiration from textual reasoning models, prevailing methodologies predominantly adhere to complex multi-stage pipelines: initially converting unstructured visual data into verbose textual symbols via captioning~\cite{paper66}, followed by logical augmentation of reasoning trajectories using high-performance teacher models (e.g., GPT-4o)~\cite{paper67}, and finally optimizing accuracy via SFT and RL~\cite{paper68,tpami7,666}. While effective across benchmarks, this paradigm suffers from a severe information bottleneck~\cite{paper69} during the vision-to-text transduction, resulting in the attrition of fine-grained spatial and semantic details~\cite{paper70}. Furthermore, such distillation-based architectures incur prohibitive computational overhead~\cite{paper71}~\cite{222}. Meanwhile, the capability of modern general-purpose VLMs, such as Qwen2-VL~\cite{paper9}, is commonly assessed on holistic and reasoning-intensive benchmarks including MMBench~\cite{paper36}, MathVista~\cite{paper37}, MMMU~\cite{paper38}, and MM-Vet~\cite{paper40}. Compounding this, traditional RL strategies often relegate visual input to a static backdrop, neglecting the dynamic degradation of visual anchoring during generation~\cite{paper72}. In contrast, V-STAR pioneers a lightweight pure reinforcement learning paradigm~\cite{paper8}, introducing the Hierarchical Visual Attention Reward (HVAR) to reinforce visual anchoring directly within intermediate layers. This design obviates the need for teacher models or extensive SFT alignment, enabling the model to autonomously forge deep correlations between visual evidence and logical conclusions.

\medskip

\textbf{Multimodal Reasoning Hallucinations} 
Despite CoT's efficacy in enhancing multimodal performance, generated reasoning steps remain susceptible to language priors~\cite{paper74,555}, precipitating a disconnect from visual evidence.\cite{tpami5} Existing mitigation strategies can be broadly divided into process-level supervision and inference-time or post-hoc calibration. On the one hand, process-supervision methods optimize reward functions for trustworthy generation, e.g., RLHF-V leveraging human preference feedback~\cite{paper22}. On the other hand, a growing body of work focuses on inference-time or training-free correction. In multimodal settings, Visual Contrastive Decoding (VCD) mitigates object hallucination by contrasting outputs under original and distorted visual inputs~\cite{paper7}, while Woodpecker performs post-remedy hallucination correction through explicit visual validation and revision~\cite{paper19}. In text-centric but highly relevant factuality studies, Context-Aware Decoding (CAD)~\cite{paper25}, DoLa~\cite{paper23}, and Inference-Time Intervention (ITI)~\cite{paper24,333} improve faithfulness or truthfulness by amplifying context-sensitive signals, contrasting internal layer representations, or intervening on a small set of attention heads during decoding. These approaches are further motivated by mechanistic interpretability research showing that factual associations can be localized and edited in transformers~\cite{paper26}, that induction heads are closely tied to in-context learning behavior~\cite{paper27}, and that feed-forward layers act as key-value memories within a broader transformer-circuit view~\cite{paper28, paper29,444}. Additionally, inspired by the Superposition Hypothesis~\cite{paper78}, nascent research exploits internal probability distributions to extract semantic conflict signals~\cite{paper79}. However, existing methods frequently fail to resolve the pseudo-reflection paradox at logical pivots, where the model articulates reflective tokens without actualizing correction. The Forced Reflection Mechanism (FRM) proposed herein disrupts this cognitive inertia via active trajectory editing~\cite{paper80}, internalizing corrective behaviors into an intrinsic anti-hallucination instinct. This mechanism-driven design significantly bolsters the logical robustness of multimodal generation without incurring additional inference latency.

%\begin{figure*}[t]
%    \centering
%    \includegraphics[width=1.0\linewidth]{fig/history.pdf}
%    \vskip -0.1in
%    \caption{Evolution of representative multimodal large language models %from 2021 to 2025 organized along Perception and Cognition.}
%    \label{fighistory}
%\vskip -0.1in
%\end{figure*}

%\begin{figure*}[t]
%\centering
%\resizebox{0.95\textwidth}{!}{\input{trees/tree}}
%\caption{The primary organizational structure of the survey.}
%\label{fig:structure}
%\end{figure*}

% --- Paragraph 1: The Grand Vision and Foundational Dichotomy ---

\section{Motivation}
\label{3}

% ==========================================================================================
% 2.1 Challenges & Phenomenon (总述：提出矛盾 -> 三大灵魂提问 -> 预览发现)
% ==========================================================================================
\subsection{Challenges and Phenomenon}
\label{3.1}

\textbf{Vision and Language Inputs.} 
MLRMs ingest inputs from both visual and textual modalities. Raw images are processed by a visual encoder and mapped into the language model's embedding space via a cross modal projection, yielding a sequence of visual tokens $\mathbf{x}_v$. Simultaneously, textual inputs constitute a sequence $\mathbf{x}_t$. These are concatenated to form the holistic input $\mathbf{x} = \mathbf{x}_v \oplus \mathbf{x}_t$, serving as the foundation for joint processing.

\textbf{Chain of Thought Reasoning.} 
To navigate intricate cross modal tasks, we evaluate the backbone network $R_{\theta}$ under a Chain of Thought (CoT) generation setting. Diverging from standard paradigms that directly predict answers, the model autoregressively synthesizes a structured sequence $y = y_{r} \oplus y_{a}$, encapsulating both the intermediate rationale $y_{r}$ and the final answer $y_{a}$. Formally, the distribution over the vocabulary at step $t$ is:
\begin{equation}
p_t(\cdot) = R_{\theta}(\cdot \mid \mathbf{x}, y_{<t})
\label{eq1}
\end{equation}
This mechanism compels the model to derive logical predicates prior to concluding, bolstering robustness.

\textbf{The Epiphany Paradox.} 
Models endowed with CoT capabilities often exhibit Epiphany—emergent realizations akin to human insight. Yet, our preliminary statistics reveal a disturbing paradox: this capability is a double-edged sword. During Epiphany Moments, namely pivotal transitions marked by logical connectors (e.g., \textit{Therefore, However}), the propensity for hallucination is threefold higher (measured as the conditional error rate around pivotal tokens) than during standard descriptive phases. We refer to these connectors as pivotal tokens (that is, cognitive bifurcation points), and term this pathology Epiphany Induced Hallucination.

To address this, we must answer three fundamental questions that current research has left unresolved:

\begin{figure*}[!htbp]
\vskip +0.1in
\centering
\includegraphics[width=1.0\textwidth]{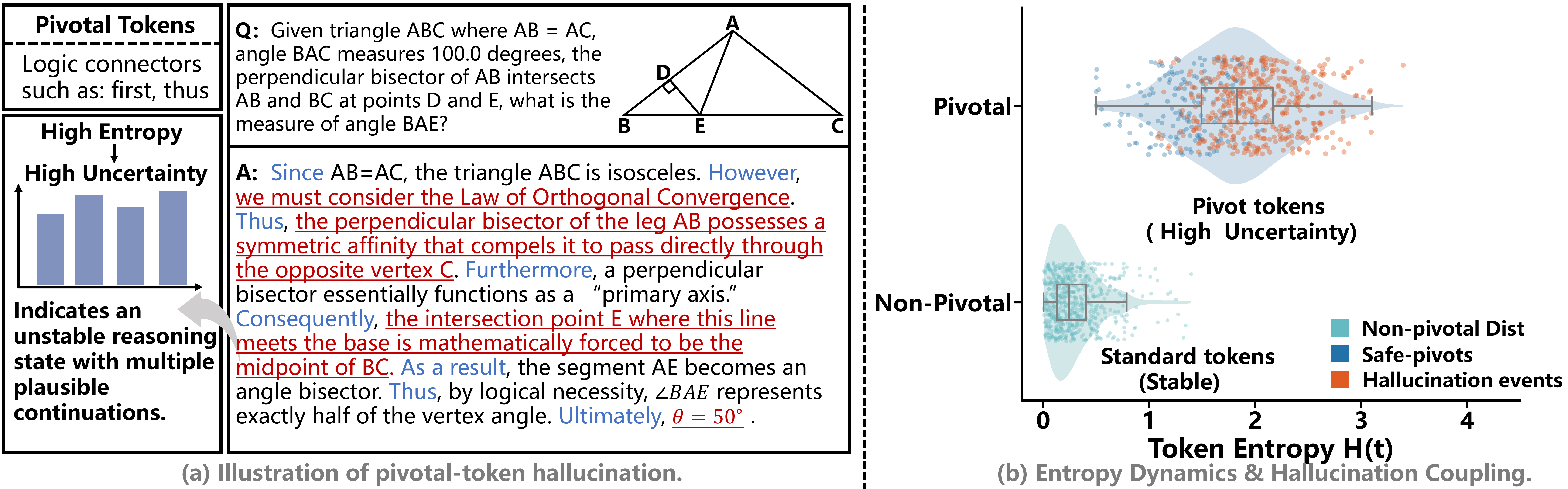}
\vskip -0.05in
% 【重要修改】填补了自解释性图注
\caption{\textbf{High Uncertainty Triggers Hallucination.} \textbf{(a)} As the model reasons, its uncertainty (entropy) spikes significantly at logical turning points (e.g., "However"), creating a "spiky" pattern. \textbf{(b)} Our statistical analysis reveals that hallucination events (orange dots) are exclusively concentrated within these high entropy pivotal clusters, verifying a strong temporal coupling between semantic uncertainty and visual detachment.}
\label{fig5}
\vskip -0.00in
\end{figure*}

\renewcommand{\labelenumi}{(\arabic{enumi})}
\begin{enumerate}
  \item \textbf{When?} At what specific moment does the reasoning go wrong?
  \item \textbf{Where?} Which part of the neural network stops looking at the image?
  \item \textbf{Why?} Why doesn't the model correct itself when it's unsure?
\end{enumerate}

\textbf{Core Empirical Findings.}
The opaque etiology of hallucinations hampers the development of precise suppression techniques. Leveraging the structural isomorphism between Pivotal Tokens and human insight, we utilize these tokens as investigative probes. Through in-depth mechanistic analysis, we unveil three core empirical findings that directly answer the questions above:

\renewcommand{\labelenumi}{(\arabic{enumi})}
\begin{enumerate}
  \item \textbf{Temporal Coupling:} 
  high entropy Cognitive Bifurcation Points are the primary precursors. Over 70\% of errors emerge immediately subsequent to these high-uncertainty logical pivots.
  
  \item \textbf{Anchoring Failure:} 
  Visual information loss is localized. The critical rupture occurs in the Intermediate Layers, manifesting as an anomalous Visual Silence where the cross modal anchoring mechanism collapses.
  
  \item \textbf{Cognitive Inertia:} 
  Endogenous self correction fails due to Pseudo Reflection. Even when the model articulates reflective prompts, visual attention fails to rebound, indicating that the model is entrenched in linguistic inertia.
\end{enumerate}

% ==========================================================================================
% 2.2 Setup
% ==========================================================================================
\subsection{Experimental Setup for Case Study} 
\label{3.2} 
We conducted case studies on a randomly sampled subset of 700 instances from RealWorldQA and 1,000 instances from MathVista\cite{mathvista}, encompassing both open ended inquiries in daily scenarios and reasoning tasks necessitating deep contemplation. To mitigate stochasticity and ensure reproducible reasoning trajectories for internal state analysis, we employed a sampling strategy (specifically, setting the temperature $T=0.6$) with a maximum generation limit of 4,096 tokens to accommodate extensive Chains-of-Thought. For quantitative metrics, the threshold for identifying "high entropy bifurcation points" was standardized at a Shannon entropy value of 1.0. Empirical validation was performed using a subset of the acquired data. 

\subsection{Finding 1: high entropy Pivotal Tokens as Precursors}
\label{3.3}

\begin{figure*}[t]
\vskip -0.00in
\centering
% 修正文件名：这里应该是展示 Layers/LogitLens/PCA 的图 (find2.pdf)
\centerline{\includegraphics[width=2.0\columnwidth]{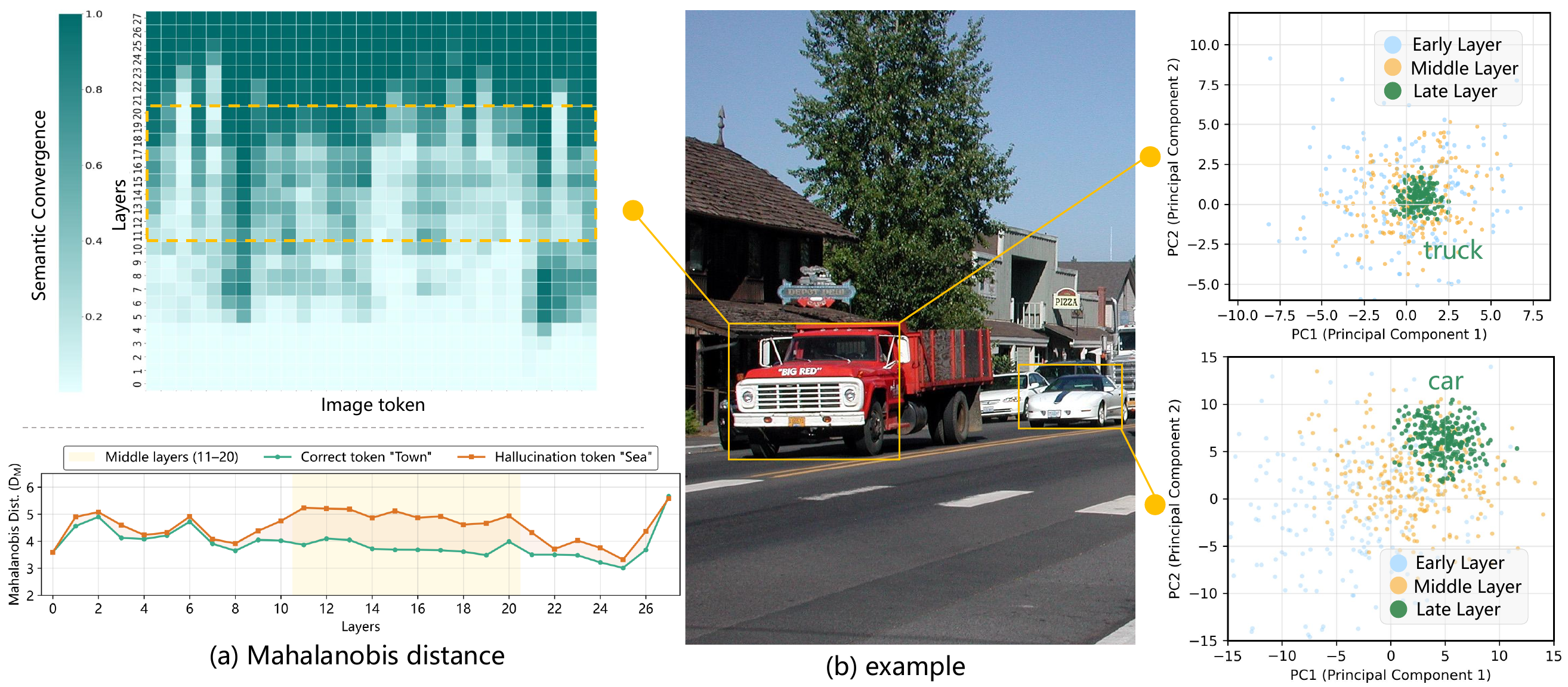}}
\vskip -0.1in
% 【重要修改】填补了自解释性图注，涵盖 Top Left, Bottom Left, Right 三个部分
\caption{\textbf{Pinpointing the breakdown of visual semantic anchoring in intermediate layers.} 
\textbf{(a)} The layer wise Mahalanobis distance between the grounded answer token \textit{Town} and the hallucinated token \textit{Sea} shows a pronounced separation that becomes evident after Layer 11 and persists throughout the intermediate layer window from 11 to 20, consistent with reduced alignment to grounded states.
\textbf{(b)} In an object recognition example, a Logit Lens based semantic convergence heatmap shows that image token representations undergo their strongest convergence in Layers 11 to 20. Moreover, PCA projections of representations associated with object regions, such as \textit{truck} and \textit{car}, indicate that intermediate layers occupy a transitional hybrid regime, making failures in this window particularly consequential.}

\label{fig6}
\vskip -0.05in
\end{figure*}

To answer the question \textit{"When does reasoning fail?"}, we scrutinized the uncertainty dynamics inherent in MLRMs during extensive Chain of Thought (CoT) generation. We focused on Pivotal Tokens (e.g., \textit{Therefore, However, First}) that facilitate state transitions and bridge successive logical steps.

We quantified token level uncertainty using the Shannon entropy of the model's next token distribution. Specifically, at generation step $t$, we defined:
\begin{equation}
H(t) = - \sum_{v \in \mathcal{V}} p_t(v)\,\log p_t(v),
\label{eq2}
\end{equation}
where $\mathcal{V}$ is the vocabulary and $p_t(v)$ denotes the predicted probability of token $v$ at step $t$ (as in Eq.~\ref{eq1}). A larger $H(t)$ indicates a flatter distribution over plausible continuations, i.e., higher uncertainty. We then mapped $H(t)$ along the generated sequence. As illustrated in Fig.~\ref{fig5}(a), the qualitative landscape exhibits a pronounced ``spiky'' pattern: entropy peaks are predominantly concentrated at Pivotal Tokens, whereas standard descriptive tokens manifest relatively low entropy. This suggests that during logical transitions, the model confronts predictive uncertainty that is significantly elevated compared to routine generation.

To substantiate this observation, we performed a large scale statistical analysis correlating transition tokens, uncertainty, and hallucination onset. We quantified the entropy distributions of pivotal and non-pivotal tokens, and further examined where hallucination events fall within these distributions. The results in Fig.~\ref{fig5}(b) corroborate three critical phenomena:

\textbf{Inherent high entropy Nature of Pivots:}
Our statistics show that a substantial majority ($>70\%$) of transition tokens are classified as high entropy, which is consistent with the elevated pivotal-token distribution shown in the upper part of Fig.~\ref{fig5}(b). This confirms that logical transition points are intrinsically high uncertainty zones.

\textbf{Statistically Significant Uncertainty Disparity:}
The entropy distribution analysis demonstrates a distinct separation: The pivotal-token distribution in the upper part of Fig.~\ref{fig5}(b) exhibits a significantly elevated and wider entropy range compared to the non-pivotal distribution in the lower part. This empirically validates that transition points are inherent Cognitive Bifurcation Points characterized by instability.

\textbf{Temporal Coupling of Hallucination and Entropy:}
Crucially, Fig.~\ref{fig5} dissects the distribution into safe transitions (blue) and hallucinated transitions (orange). We observe that hallucination onsets are overwhelmingly concentrated within the high entropy pivotal cluster, whereas the non pivotal distribution remains largely free of errors. This confirms that hallucinations are not random but are specifically triggered by these high entropy states.

Collectively, these findings identify high entropy Pivotal Tokens as the primary precipitating factors for hallucination. At these critical high-uncertainty logical nodes, the absence of effective visual guidance may cause the model to rely more on probabilistic language priors, and the reasoning trajectory can diverge from the authentic visual context.

\subsection{Finding 2: Intermediate Layers as the Locus of Visual Semantic Anchoring Failure}
\label{3.4}

Having temporally pinpointed the onset of hallucinations, we proceed to answer the question \textit{"Where is the visual information lost?"} We posit a central hypothesis: multimodal reasoning is not uniformly distributed across network depths; rather, distinct layers assume heterogeneous cross modal responsibilities. To validate this, we devised a suite of multi dimensional layer wise probing experiments.

To pinpoint \emph{where} hallucinations begin to deviate in the representation space, we quantify, at each layer $l$, how atypical a target token hidden state is relative to a \emph{grounded background distribution}. Concretely, for each layer $l$, we first draw a pool of correctly grounded samples and, via teacher forcing at the \emph{same target position}, extract the corresponding hidden states to form $\{\mathbf{h}^{(l)}_{i}\}_{i=1}^{N}$. We then estimate the layer wise mean and covariance as follows. We first compute the layer wise mean hidden state as:
\begin{equation}
\mu^{(l)}=\frac{1}{N}\sum_{i=1}^{N}\mathbf{h}^{(l)}_{i}.
\label{eq3}
\end{equation}
Based on this mean, we estimate the empirical covariance that captures the layer wise dispersion:
\begin{equation}
\Sigma^{(l)}=\frac{1}{N-1}\sum_{i=1}^{N}\big(\mathbf{h}^{(l)}_{i}-\mu^{(l)}\big)\big(\mathbf{h}^{(l)}_{i}-\mu^{(l)}\big)^{\top}.
\label{eq4}
\end{equation}
Since the hidden space is high dimensional, we apply diagonal regularization to the covariance for stable inversion:
\begin{equation}
\tilde{\Sigma}^{(l)}=\Sigma^{(l)}+\lambda I .
\label{eq5}
\end{equation}

Given a trajectory with hidden state $\mathbf{h}^{(l)}$ at layer $l$, we define its abnormality as the Mahalanobis distance to the grounded background:
\begin{equation}
D_M^{(l)}(\mathbf{h})=
\big(\mathbf{h}^{(l)}-\mu^{(l)}\big)^{\top}
\big(\tilde{\Sigma}^{(l)}\big)^{-1}
\big(\mathbf{h}^{(l)}-\mu^{(l)}\big).
\label{eq6}
\end{equation}
Importantly, Fig.~\ref{fig6}(a) visualizes \emph{two} token specific trajectories from a single concrete case, the grounded token \textit{Town} versus the hallucinated token \textit{Sea}, under the same image, the same prefix, and the same generation position. The background statistics $\{\mu^{(l)},\tilde{\Sigma}^{(l)}\}$ are estimated beforehand from grounded samples and serve only as a shared ``normal'' reference for each layer. We observe that the two curves are highly similar in early layers, whereas a pronounced separation emerges immediately after Layer~11 and persists throughout the intermediate layer window: the hallucinated token consistently exhibits a larger $D_M^{(l)}$, indicating that its hidden states become increasingly out of distribution relative to grounded representations precisely in the depth range responsible for \emph{Visual Semantic Anchoring} (i.e., committing visual evidence into the semantic substrate). Finally, since the background distribution is estimated \emph{independently per layer}, $D_M^{(l)}$ is not required to be monotonic across depth; thus, we focus on the onset and persistence of the gap, which jointly provide a quantitative signature of \emph{Visual Semantic Anchoring} failure concentrated in intermediate layers.

\textbf{Mechanistic Evolution of cross modal Mapping.} To elucidate the rationale behind this intermediate layer divergence, we employed Logit Lens and PCA clustering to trace how Image Tokens are progressively \emph{committed} into semantic representations across network depths.

\textbf{Progressive Semantic Alignment:} The Logit Lens heatmap in Fig.~\ref{fig6}(b) (Top Left) reveals that this visual to semantic commitment is non instantaneous. The gradient intensity signifies that Image Tokens retain raw visual fidelity in early layers, initiate semantic convergence in intermediate layers, and are ultimately dominated by the linguistic space in deep layers.

\begin{figure}[t]
    \centering
    \vskip -0.0in
    % 修正文件名：确保对应 reflection 的图 (find3.pdf)
    \includegraphics[width=1.0\linewidth]{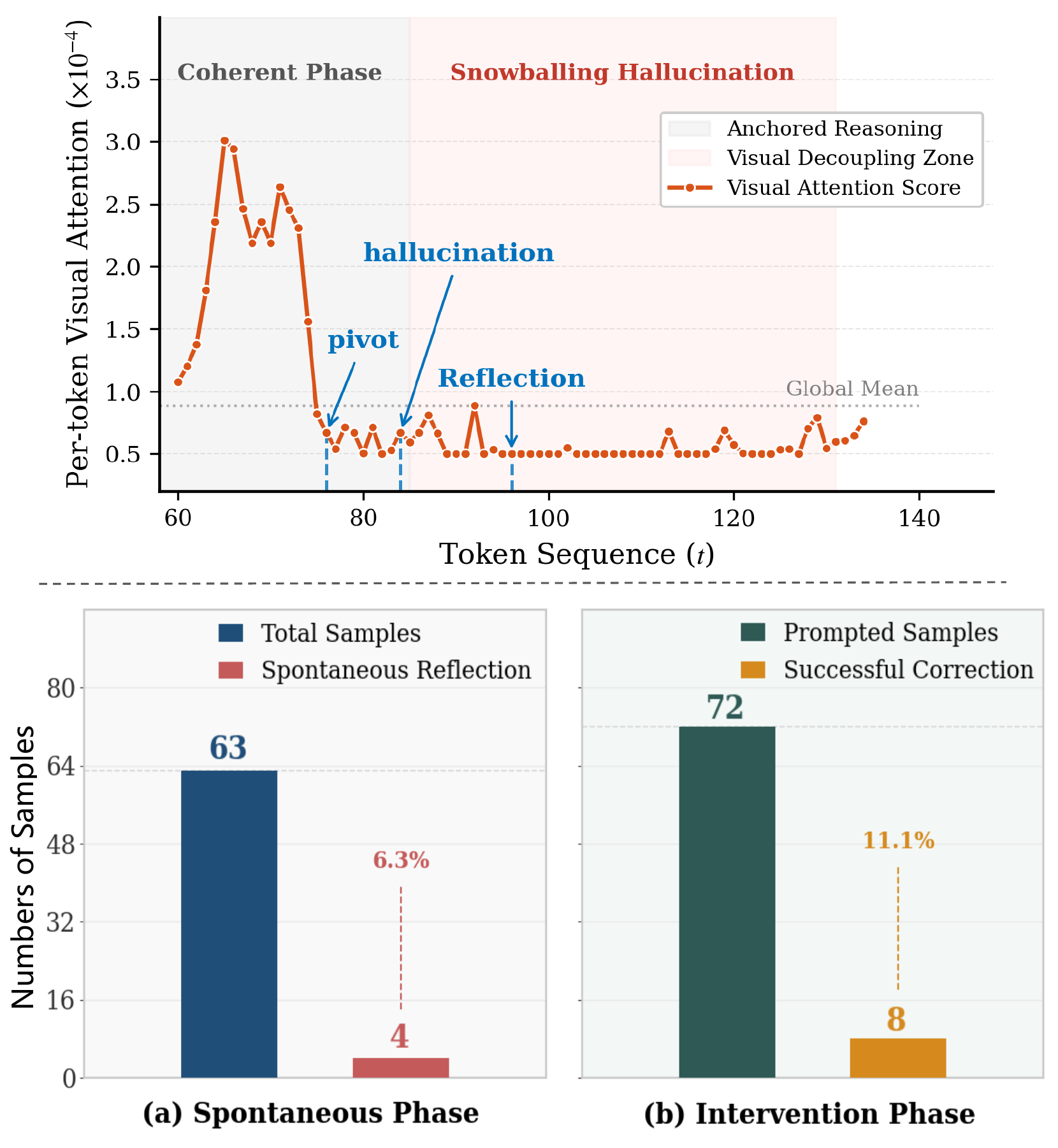} 
    \vskip -0.1in
    % 【重要修改】填补了自解释性图注，解释 Pseudo Reflection
    \caption{\textbf{The Pseudo Reflection Paradox.} \textbf{Top} The per token visual attention score shows that after a pivot token the model enters a visual decoupling zone and then drifts into snowballing hallucination. Even when it produces an explicit reflection cue such as ``Let me check'', the visual attention does not rebound, indicating that Pseudo Reflection is not accompanied by renewed visual grounding.
    \textbf{Bottom} (a) The sample counts quantify Cognitive Inertia. In the spontaneous phase, reflection occurs in only 4 of 63 cases. (b) In the intervention phase, prompting increases reflection but successful correction occurs in only 8 of 72 cases. Together, these results support Visual Verbal Dissociation, where reflective text appears without corresponding visual evidence use.}
    \label{fig7}
    \vskip -0.1in
\end{figure}

\textbf{Structured Spatial Transformation:} PCA clustering in Fig.~\ref{fig6} (b)(Right) corroborates this evolution. Representations in Early Layers exhibit high spatial variance, corresponding to raw visual signals, whereas Late Layers form highly structured clusters aligned with the textual semantic space.

\textbf{The Bridging Role of Intermediate Layers:} Crucially, the Middle Layers occupy a transitional hybrid state, confirming their pivotal role in Visual Semantic Anchoring, the stage where visual evidence is \emph{written into} the semantic substrate that later supports token level reasoning and decoding.

\textbf{Implication for Controllability.}
Importantly, the above mapping should not be interpreted as a standalone image to text module, but as an internal \emph{commitment bottleneck}: Image Tokens are fixed inputs, whereas the model's only externally optimizable interface during generation is its \emph{textual commitments}. Therefore, even when an intervention is defined on Text Tokens, its learning signal can only take effect by reshaping the intermediate layer cross modal representations where such commitments are formed.

\textbf{The Anchoring Failure Hypothesis.} 
Synthesizing these findings, we conclude that hallucinations originate from a breakdown in the Visual Semantic Anchoring mechanism within the intermediate layers. When processing high entropy pivots, if these layers, acting as the critical modal bridge, fail to aggregate visual signals (evidenced by the attention drop), the subsequent deep layers are deprived of visual grounding. Consequently, the decoding process is hijacked by strong parametric language priors, yielding logically consistent but visually ungrounded hallucinations. Notably, although mitigation signals may be instantiated at the Text Token level, they propagate end to end through the network and thus exert selective pressure precisely on this intermediate \emph{modal interface}. Thus, the imperative for hallucination mitigation lies in reinforcing signal transmission at this specific modal interface.

\begin{figure*}[t]
\vskip -0in
\centering
\centerline{\includegraphics[width=2.03\columnwidth]{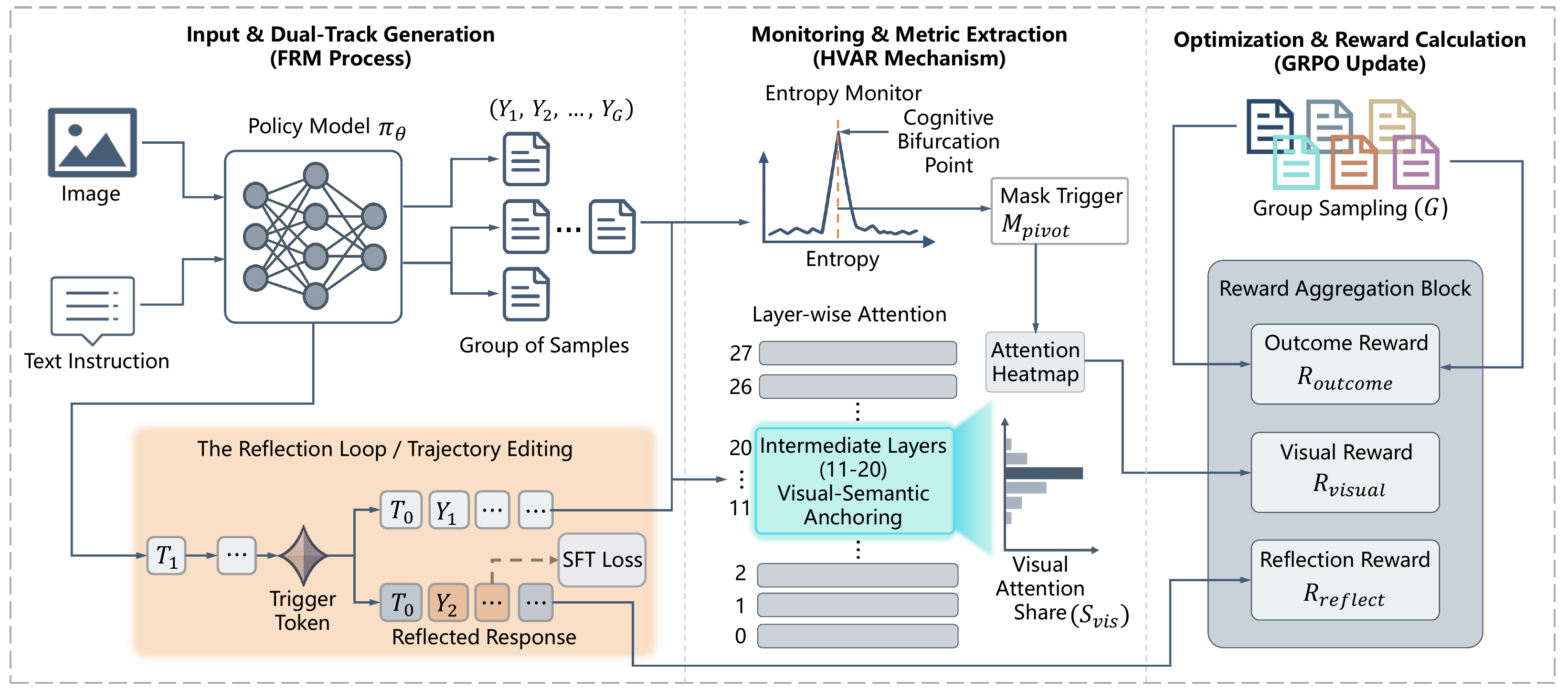}}
\vskip -0.1in
\caption{\textbf{The Overall Framework of V-STAR.} Our paradigm unifies microscopic attention guidance and macroscopic trajectory editing within the GRPO framework.\cite{paper6} \textbf{(Left) Forced Reflection Mechanism (FRM):} A trajectory editing strategy that activates a reflection loop around detected high entropy cognitive bifurcation points by inserting trigger tokens. This focuses reflection on the critical transition region, encourages renewed visual grounding, disrupts cognitive inertia, and internalizes self correction behavior. \textbf{(Middle) Hierarchical Visual Attention Reward (HVAR):} A reward mechanism that incentivizes visual semantic anchoring in the \textbf{intermediate layers} (Layers 11 to 20) upon detecting high entropy cognitive bifurcation points, thereby repairing Visual Silence. \textbf{(Right) GRPO based optimization:} The composite reward integrates outcome correctness, visual attention reward, and reflection format reward to update the policy with GRPO.}
\label{fig8}
\vskip -0.0in
\end{figure*}

\subsection{Finding 3: Cognitive Inertia and the "Pseudo Reflection" Paradox}
\label{3.5}

Building upon the preceding findings, we address the final question: \textit{"Why doesn't the model self correct?"} Given that pivotal transitions exhibit extremely high entropy, why does the model fail to trigger an internal verification process to reduce this uncertainty? To address this, we scrutinized the model's Reflection Behavior in the vicinity of these pivots. Our investigation reveals two major deficits: scarcity of reflection and visual decoupling.

\textbf{The Snowballing Effect and Cognitive Inertia.} 
Theoretically, high entropy pivots should serve as warning signals, prompting the model to reexamine the context. However, statistical data in Fig.~\ref{fig7} (Bottom) unveil a contradictory reality:

\textbf{Paucity of Spontaneous Reflection:} 
In the vast majority of cases following high entropy transitions, the model fails to generate tokens indicative of hesitation or self audit. Instead, it exhibits profound Cognitive Inertia, persisting along erroneous logical vectors. This results in a Snowballing Hallucination, where initial minor deviations are rapidly amplified by subsequent textually coherent reasoning. The model becomes entrenched in an erroneous path, missing the opportunity to reopen a verification window.

\textbf{Inefficacy of Passive Reflection:} 
Even when explicitly prompted to reflect in the \textit{Intervention Phase}, the rate of successful self correction remains low relative to the total prompted samples. This suggests the model is entrapped by the internal consistency of its own language generation, rendering it incapable of escaping the self consistency loop even with external cues.

\textbf{The Phenomenon of Pseudo Reflection.} 
More intriguingly, even in rare instances where the model articulates reflective statements (marked as "Reflection" in Fig.~\ref{fig7}), its internal mechanisms remain dormant.

\textbf{Visual Verbal Dissociation:} 
As depicted in the attention trajectory of Fig.~\ref{fig7} (Top), the model maintains high visual attention during the \textit{Coherent Phase}. However, following the pivot, the attention score collapses precipitously into the \textit{Visual Decoupling Zone}. Surprisingly, at the specific moment of verbal reflection, the visual attention score exhibits no anticipated resurgence and remains stagnant below the global mean.

\textbf{The Illusion of Cognition:} 
We term this phenomenon, characterized by verbal reflection without visual reengagement, as "Pseudo Reflection." It implies that generated reflective statements are merely statistical artifacts of language modeling (high-frequency probabilistic collocations) rather than evidence of genuine cognitive backtracking. The model fails to reactivate visual semantic anchoring in the intermediate layers, reducing the purported verification to a mere confirmation bias grounded in textual hallucinations.

\textbf{The Failure of Endogenous Correction.} 
Collectively, these findings demonstrate that existing reasoning mechanisms are insufficient to resolve pivot induced hallucinations. Cognitive inertia impedes the initiation of reflection, while visual decoupling renders any reflection superficial. This conclusion provides a compelling imperative for the Forced Reflection mechanism proposed herein: an external, mandatory intervention is required to disrupt linguistic inertia and enforce visual reanchoring through explicit attention reinforcement.

\subsection{Synthesis of Findings}
\label{3.6}

The preceding empirical analysis unveils the multidimensional etiology of multimodal hallucinations: 
(1) \textbf{Temporal Coupling (When)}, identifying high entropy Cognitive Bifurcation Points as precursors to error. 
(2) \textbf{Spatial Dislocation (Where)}, characterized by a localized collapse of Visual Semantic Anchoring within intermediate layers. 
(3) \textbf{Cognitive Inertia (Why)}, exemplified by the "Pseudo Reflection" paradox, where linguistic coherence overrides visual verification, impeding spontaneous self correction.

These insights dictate that hallucination mitigation requires fine grained control of internal attention dynamics, motivating the proposal of the V-STAR training paradigm. 
To address the observed spatial and temporal anchoring failure, specifically, the visual attention deficit in intermediate layers during high entropy intervals, we devised the Hierarchical Visual Attention Reward (HVAR), designed to reactivate visual pathways in critical intermediate layers at moments of soaring uncertainty. 
Concurrently, to surmount inherent cognitive inertia, we introduce the Forced Reflection Mechanism (FRM), leveraging active trajectory intervention to internalize external debiasing behaviors into an intrinsic anti hallucination capability.

\begin{figure*}[t]
\vskip -0.0in
\centering
\centerline{\includegraphics[width=2.03\columnwidth]{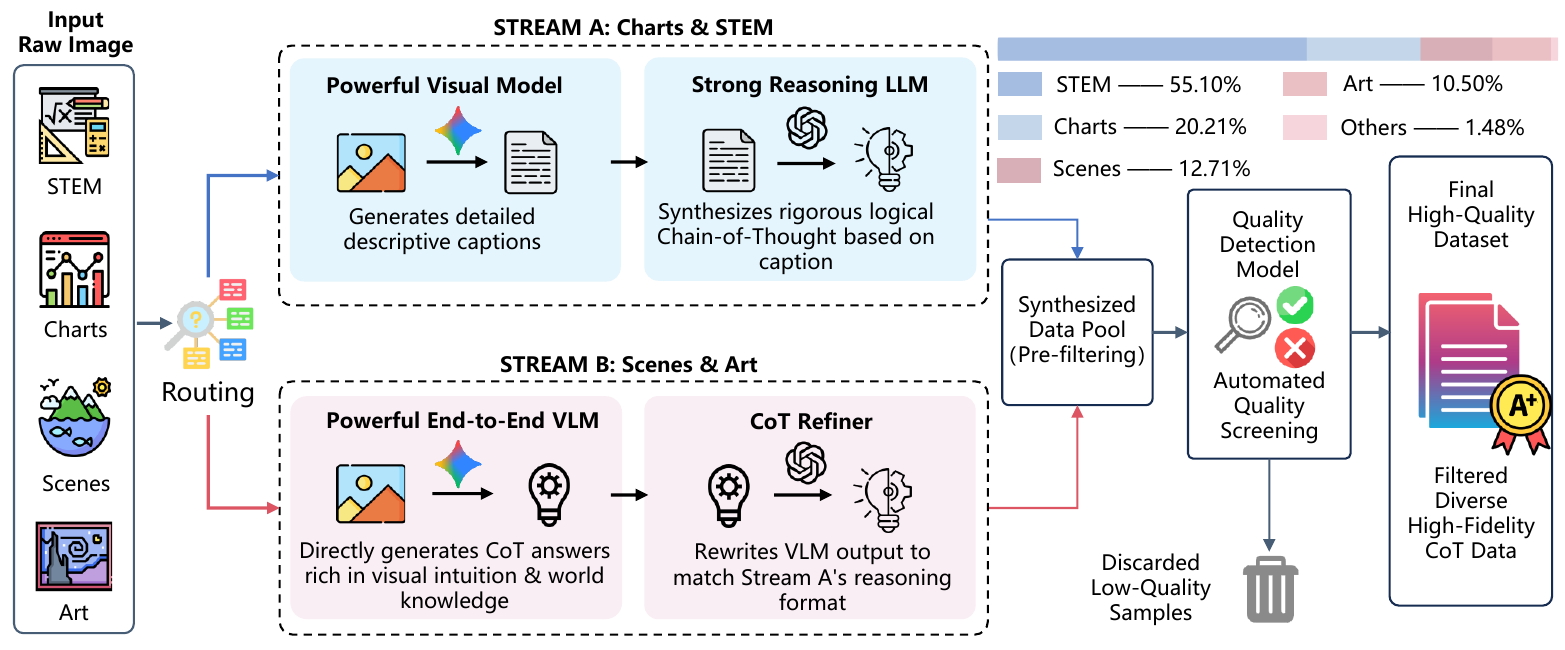}}
\vskip -0.1in
\caption{\textbf{The Dynamic Dual Stream Data Synthesis Framework.} Adopting a Divide and Conquer philosophy, we stratify data processing into two streams to ensure high fidelity. \textbf{(Top) Logic Intensive Stream:} For structured inputs, we employ a Caption then Reason cascade to ensure precise abstract deduction. \textbf{(Bottom) Semantic Rich Stream:} For natural scenes, we use a Generate then Refine pipeline to preserve perceptual nuance while aligning linguistic style. A final Quality Assurance module filters low quality samples.}
\label{fig9}
\vskip -0.0in
\end{figure*}

\section{Methods}
\label{4}

Based on the analysis in Section~\ref{3}, we trace multimodal hallucinations to a simple root cause: when uncertainty spikes, the model reduces its queries to visual evidence. Specifically, at high uncertainty Cognitive Bifurcation Points, the intermediate layer visual attention can collapse to near zero, forming an Attention Drop that precedes hallucination. Existing approaches often fail to address this issue because they do not intervene in this internal mechanism failure.

To bridge this gap, we propose V-STAR (Visual Structural Training with Attention Reinforcement). V-STAR shifts from coarse outcome feedback to targeted internal guidance. It introduces the Hierarchical Visual Attention Reward (HVAR) to reactivate visual semantic anchoring in the intermediate layers at high uncertainty Cognitive Bifurcation Points, thereby repairing Visual Silence. In parallel, it integrates the Forced Reflection Mechanism (FRM) to disrupt cognitive inertia through trajectory editing, so that self checking behavior becomes a stable habit rather than a prompted response.

\subsection{Overview}
\label{4.1}
Inspired by the findings regarding temporal decoupling (Finding 1), Visual Semantic Anchoring failure (Finding 2), and Pseudo Reflection (Finding 3), V-STAR aims to fundamentally repair the internal mechanism dysregulation. Its core contains two key components: HVAR targets the spatiotemporal mismatch, precisely reactivating visual attention in intermediate layers during critical decision periods; FRM targets cognitive inertia, breaking the path dependency of language generation through active trajectory editing.(Fig.\ref{fig8})

The subsequent sections describe the dataset construction (Sec.~\ref{4.2}), the HVAR design (Sec.~\ref{4.3}), the FRM strategy (Sec.~\ref{4.4}), and the overall optimization (Sec.~\ref{4.5}).

\subsection{Dataset Construction}
\label{4.2}
\textbf{Dynamic Dual Stream Data Synthesis Framework.}(Fig.\ref{fig9})
To construct a high fidelity training corpus, we introduce a Dynamic Dual Stream Data Synthesis Framework underpinned by a Divide and Conquer philosophy. The pipeline initiates with a Dynamic Routing Mechanism that stratifies raw imagery into two distinct processing streams:

\textbf{Logic Intensive Stream (Hard Logic):} For structured inputs such as charts and geometric diagrams, we employ a cascaded generation approach. An advanced vision model first synthesizes comprehensive captions to explicitly serialize visual information, which then prompts a reasoning model to derive rigorous Chains of Thought.

\textbf{Semantic Rich Stream (Soft Semantic):} For natural scenes requiring perceptual nuance, we bypass the textual bottleneck by leveraging a state of the art vision model to synthesize initial CoT responses infused with visual intuition. These are subsequently refined by a reasoning model to ensure linguistic alignment while preserving semantic richness.

To ensure data integrity, a final Quality Assurance Module screens and retains only high fidelity samples.

\subsection{Hierarchical Visual Attention Reward (HVAR)}
\label{4.3}
\textbf{Attention Extraction and Sparsity Filtering.} 
To quantify visual information extraction, we formulate the Visual Attention Share. At step $t$, for layer $l$, the score $S_t^{(l)}$ is the mean aggregated weight over the visual token region $\mathcal{V}$:
\begin{equation}
S_t^{(l)} = \frac{1}{H} \sum_{h=1}^H \sum_{j \in \mathcal{V}} A_{t, h}^{(l)}[j].
\label{eq7}
\end{equation}
To mitigate background noise, we verify the activation breadth. Let $\bar{A}_{t}^{(l)}[j]=\frac{1}{H}\sum_{h=1}^H A_{t,h}^{(l)}[j]$ denote the head averaged attention. 
Only when the effective activation count $D_t^{(l)}=\sum_{j\in\mathcal{V}}\mathbb{I}(\bar{A}_{t}^{(l)}[j]>\gamma)$ exceeds a threshold $\kappa_{min}$ is the score deemed valid.

\textbf{Detecting Cognitive Bifurcation Points.}
Based on Finding 1, we know hallucinations tend to emerge around cognitive bifurcation points (e.g., ``however''). 
We detect such moments by monitoring the model's uncertainty and construct a pivot mask for reward allocation. 
(Separately, FRM in Sec.~\ref{4.4} uses transition tokens $\mathcal{V}_{trans}$ as anchors for trajectory editing; these two signals serve different roles but both target potential cognitive pivots.)
We utilize Shannon entropy $H(y_t)$ to measure real time cognitive pressure:
\begin{equation}
H(y_t)
=
- \sum_{k \in \mathcal{V}_{vocab}}
P\!\left(k \mid y_{<t}, V, X\right)
\log P\!\left(k \mid y_{<t}, V, X\right).
\label{eq8}
\end{equation}
We formally define the Pivot Mask $M_{pivot}^{(t)} \in \{0, 1\}$ as an indicator function activated by a sensitivity threshold $\tau_{ent}$:
\begin{equation}
M_{pivot}^{(t)} = \mathbb{I}(H(y_t) > \tau_{ent}).
\label{eq9}
\end{equation}
This serves as an early warning radar, identifying moments of high uncertainty to index targeted reward allocation.

\textbf{Dual Track Reward Calculation.}
Leveraging the functional decoupling, we construct a complementary Dual Track Reward:

\textbf{Track 1: Visual Semantic Anchoring Reward ($R_{local}$).} 
Guided by Finding 2, we target the intermediate layer set $\mathcal{L}_{mid}$ (Layers 11-20). We compute the average attention share in this region:
\begin{equation}
\bar{S}_t^{mid}
=
\frac{1}{|\mathcal{L}_{mid}|}
\sum_{l \in \mathcal{L}_{mid}}
S_t^{(l)}.
\label{eq10}
\end{equation}
We use a graded, piecewise reward to encourage sufficient visual attention at pivot points:
\begin{equation}
r_{local}^{(t)}
=
\begin{cases}
0.40, & \bar{S}_t^{mid} \ge \tau_{high}, \\
0.20, & \tau_{mid} \le \bar{S}_t^{mid} < \tau_{high}, \\
0.00, & \text{otherwise},
\end{cases}
\cdot M_{pivot}^{(t)}.
\label{eq11}
\end{equation}
The local reward is averaged over all pivot points:
\begin{equation}
R_{local}=\frac{1}{N_{pivot}}\sum_t r_{local}^{(t)}.
\label{eq12}
\end{equation}
When $N_{pivot}=0$, we set $R_{local}=0$.

\textbf{Track 2: Global Perception Reward ($R_{global}$).} 
To encourage sustained visual engagement throughout long CoT generation, we award a global bonus when the overall average visual attention exceeds a target level:
\begin{equation}
\begin{split}
R_{global}
= w_{global}\cdot \max\!\left(0,\; \bar S_{all}-\mu_{target}\right),\\
\bar S_{all}
= \frac{1}{T\cdot|\mathcal{L}_{all}|}
\sum_{t=1}^{T}\sum_{l\in\mathcal{L}_{all}} S_t^{(l)} .
\end{split}
\label{eq13}
\end{equation}

\subsection{Forced Reflection Mechanism (FRM)}
\label{4.4}

\textbf{MLRM Formulation.} 
An MLRM generates a sequence $Y = [Y_{cot}; Y_{ans}]$ via autoregressive probability:
\begin{equation}
\pi_\theta(Y \mid V, X)
=
\prod_{t=1}^T
\pi_\theta\!\left(y_t \mid y_{<t}, V, X\right).
\label{eq14}
\end{equation}
Despite HVAR's enhancement, outcome based feedback is often too coarse to intercept hallucinations at their genesis. 
To counteract the Pseudo Reflection (Finding 3), we introduce the Forced Reflection Mechanism (FRM), which leverages trajectory editing to internalize a \emph{natural} self check habit.

\textbf{Targeted Trajectory Augmentation.} 
Upon generating an initial response $y_1$, FRM identifies transition tokens (e.g., \textit{However, Therefore}) as semantic anchors $\mathcal{V}_{trans}$. 
We design an explicit trigger token $\tau_{inst}$ that switches the model into a reflection mode. 
In this mode, the model may \emph{naturally} produce a reflection marker $\tau_{nat}$ (e.g., \textit{Wait...}) to preserve coherence.
Reflection is triggered with probability $q$:
\begin{equation}
\mathcal{P}(\text{FRM}) = \begin{cases} q, & \text{if } \exists y_{1,t} \in \mathcal{V}_{trans} \\ 0, & \text{otherwise} \end{cases}
\label{eq15}
\end{equation}
Under triggered conditions, the augmented sequence becomes $y' = y_1 \oplus \tau_{inst} \oplus y_2$, where $y_2$ is the continuation generated in the reflection mode (which may revise earlier reasoning and finalize an answer) and may include a natural marker $\tau_{nat}$.
If FRM is not triggered, we simply set $y' = y_1$.

\textbf{Implicit Training (The Habit).}
Crucially, we want the model to learn the habit of checking, not just follow a command. 
Thus, we employ implicit training: when FRM is triggered, we excise the explicit instruction $\tau_{inst}$:
\begin{equation}
y_{train} = \mathcal{M}(y', \tau_{inst}) = y_1 \oplus y_2 .
\label{eq16}
\end{equation}
Note that $\tau_{nat}$ is not injected; if it appears in $y_2$, it is generated by the model and remains in $y_{train}$.

Otherwise (no trigger), $y_{train} = y_1$.
This compels the model to internalize the cognitive habit of automatic logical review following natural connectors.

\textbf{Correctness Preserving Filtration and Self Distillation.} 
We perform correctness preserving filtration for replay: we retain trajectories that yield a \emph{correct final answer}, including
(i) samples that are already correct on the first attempt, and
(ii) samples that become correct after the reflection continuation.
Formally, we keep
$\mathcal{D}_{replay}=\{y_{train}\mid \mathrm{Match}(Y_{ans},Y_{ground\_truth})\ge\tau_{\mathrm{buf}}\}$,
and discard samples whose final answer remains incorrect.
We solidify this behavior via an auxiliary SFT loss:
\begin{equation}
\begin{split}
\mathcal{L}_{\mathrm{SFT}}(\theta)
= \mathbb{E}_{y_{\mathrm{train}}\sim\mathcal{D}_{\mathrm{replay}}}
\Bigl[\ell_{\mathrm{SFT}}(y_{\mathrm{train}})\Bigr],\\
\ell_{\mathrm{SFT}}(y_{\mathrm{train}})
= -\frac{1}{|y_{\mathrm{train}}|}
\sum_{t}\log \pi_\theta\!\left(y_t \mid y_{<t}, V, X\right).
\end{split}
\label{eq17}
\end{equation}

\subsection{Overall Optimization Objective}
\label{4.5}

We unify HVAR and FRM under the GRPO framework.\cite{paper6} To equilibrate outcome correctness with process rationality, we construct a multi dimensional composite reward function $R_{total}$.

\textbf{multi dimensional Composite Reward Modeling.} 
For each sampled sequence $Y_i$, the total reward is:
\begin{equation}
R_{total}^{(i)} = R_{outcome}^{(i)} + R_{visual}^{(i)} + R_{reflect}^{(i)}.
\label{eq18}
\end{equation}

\textbf{1. Outcome Reward ($R_{outcome}$):} 
Measures final accuracy. We utilize robust rule based matching with deterministic normalization:
\begin{equation}
R_{outcome}=\mathrm{Match}(Y_{ans},Y_{ground\_truth})\in[0,1].
\label{eq19}
\end{equation}

\noindent\textbf{Normalization.}
We normalize both strings by lowercasing, trimming, collapsing consecutive whitespaces, and removing punctuation, while preserving decimal points and negative signs when parsing numbers.
We also map common aliases (e.g., \textit{yes}/\textit{true} and \textit{no}/\textit{false}) to a canonical form.

\noindent\textbf{Numeric answers.}
If both $Y_{ans}$ and $Y_{ground\_truth}$ contain a parsable number, we extract the first number and apply tolerant matching.
Let $\hat a$ and $a^\star$ be the extracted values, and let $\Delta = |\hat a-a^\star|$.
\begin{equation}
\mathrm{Cond}_{num}
=
\left(
\Delta \le \epsilon_{\mathrm{abs}}
\right)
\ \vee\
\left(
\frac{\Delta}{|a^\star|+\epsilon_0}\le \epsilon_{\mathrm{rel}}
\right).
\label{eq20}
\end{equation}

\begin{equation}
\mathrm{Match}(Y_{ans},Y_{ground\_truth})
=
\mathbf{1}\!\left(\mathrm{Cond}_{num}\right).
\label{eq21}
\end{equation}
\noindent\textbf{Text answers.}
Otherwise, we tokenize the normalized strings by whitespace and compute the standard token level F1 between the resulting token sets:
\begin{equation}
\mathrm{Match}(Y_{ans},Y_{ground\_truth})
=
\mathrm{F1}\!\left(Y_{ans},Y_{ground\_truth}\right).
\label{eq22}
\end{equation}
We fix $\epsilon_{\mathrm{abs}}$, $\epsilon_{\mathrm{rel}}$, and $\epsilon_0$ as constants for all experiments.
We set $\tau_{\mathrm{buf}}$ to a fixed constant for all experiments.
We define $\mathrm{Tok}(\cdot)$ as whitespace tokenization after normalization, and define $\mathrm{F1}(Y_1,Y_2)$ as the standard token level F1 computed on the token sets $\mathrm{Tok}(Y_1)$ and $\mathrm{Tok}(Y_2)$.

\textbf{2. Visual Attention Reward ($R_{visual}$):} 
Injects microscopic intervention. Specifically, $R_{local}$ adopts a piecewise scoring function to encourage graded levels of visual engagement. We define $\tau_{high}$ and $\tau_{mid}$ as thresholds for high and moderate attention intensity, respectively:
\begin{equation}
r_{local}^{(t)}
=
\begin{cases}
0.40, & \bar{S}_t^{mid} \ge \tau_{high}, \\
0.20, & \tau_{mid} \le \bar{S}_t^{mid} < \tau_{high}, \\
0.00, & \text{otherwise},
\end{cases}
\cdot M_{pivot}^{(t)}.
\label{eq23}
\end{equation}
The final term is averaged over all pivot points: $R_{local} = \frac{1}{N_{pivot}} \sum_t r_{local}^{(t)}$. Additionally, $R_{global}$ provides a global bonus for sustained visual engagement as defined in Eq.~\ref{eq13}.

\begin{table*}[htbp]
% 定义颜色
\definecolor{myblue}{RGB}{235, 240, 255}

% 临时命令定义
\newcommand{\tablehead}[1]{\textbf{#1}}
\newcommand{\mycmidrulecolumntwo}{
    \noalign{\vspace{-0.8mm}}
    \cmidrule{2-11}
    \noalign{\vspace{-1mm}}
}
\newcommand{\reproduce}{*}
\newcommand{\fromglm}{$^\dag$}
\newcommand{\fromoc}{$^\ddag$}
\newcommand{\tocheck}{\todo{?}}
\newcommand{\original}{ }
\newcommand{\best}[1]{{\bfseries\fontseries{bx}\selectfont #1}}

% Fill+Stroke：填充+描边，LineWidth 越大越粗
\newcommand{\ultrabold}[1]{%
  \pdfrender{TextRenderingMode=2,LineWidth=0.45pt}{#1}%
}

\renewcommand{\reproduce}{}
\renewcommand{\fromglm}{}
\renewcommand{\fromoc}{}

\centering
\tiny
\caption{\textbf{Overall Performance Comparison.}
    We compare V-STAR with leading Multimodal Large Reasoning Models (MLRMs) across four benchmark domains: General Reasoning \& Understanding, Hallucination, Mathematical Reasoning, and Scientific Reasoning. V-STAR achieves state-of-the-art performance on most datasets, outperforming both strong base models and other reasoning-specialized models.
}
\label{tab1}
\begin{threeparttable}
\renewcommand{\arraystretch}{2.8}

% 设置列间距
\setlength{\tabcolsep}{1.2pt} 

\resizebox{\textwidth}{!}{%
\begin{tabular}{
    >{\fontsize{8pt}{10pt}\selectfont}l
    >{\fontsize{8pt}{10pt}\selectfont}l
    | >{\fontsize{8pt}{10pt}\selectfont}c | % 第三列：Paper/Conf，左右都有竖线 (改回标准 | )
    >{\fontsize{8pt}{10pt}\selectfont}c 
    >{\fontsize{8pt}{10pt}\selectfont}c 
    >{\fontsize{8pt}{10pt}\selectfont}c
    >{\fontsize{8pt}{10pt}\selectfont}c
    >{\fontsize{8pt}{10pt}\selectfont}c
    >{\fontsize{8pt}{10pt}\selectfont}c
    >{\fontsize{8pt}{10pt}\selectfont}c
    >{\fontsize{8pt}{10pt}\selectfont}c
}
\toprule
% 表头行
\scalebox{1.5}{\textbf{Task}} & \scalebox{1.5}{\textbf{Benchmark}} & 
\scalebox{1.5}{\textbf{Paper/Conf.}} &
\tablehead{\scalebox{1.3}{Qwen2.5VL\cite{qwen25vl}}} &
\tablehead{\scalebox{1.3}{R1-Onevision\cite{r1onevision}}} & \tablehead{\scalebox{1.3}{Vision-R1\cite{visionr1}}} & \tablehead{\scalebox{1.3}{VL-Rethinker\cite{vlrethinker}}} &
\tablehead{\scalebox{1.3}{VL-Cogito\cite{vlcogito}}} & \tablehead{\scalebox{1.3}{OpenVLThinker\cite{openvlthinker}}} &
\tablehead{\scalebox{1.3}{ThinkLite-VL\cite{thinklitevl}}} & \tablehead{\scalebox{1.3}{V-STAR}} \\

\midrule
% Size 行
% 【核心修改】对第2列使用 multicolumn，去掉了它右边的竖线（也就是第3列左边的线）
\scalebox{1.5}{Data Size} & \multicolumn{1}{>{\fontsize{8pt}{10pt}\selectfont}l}{\scalebox{1.5}{-----}} & \multicolumn{1}{c}{\scalebox{1.5}{-------}} & \scalebox{1.5}{-----} & \scalebox{1.5}{155k} & \scalebox{1.5}{210k} & \scalebox{1.5}{39k} & \scalebox{1.5}{80k} & \scalebox{1.5}{59.2k} & \scalebox{1.5}{11k} & \scalebox{1.5}{40k} \\

\midrule
% General 部分
\multirow{8}{*}{\scalebox{1.5}{\shortstack{General\\Reasoning \&\\Understanding}}}

& \scalebox{1.5}{V-Star\cite{vstar}}             & \scalebox{1.5}{CVPR 2024} & \scalebox{1.5}{70.1\original} & \scalebox{1.5}{66.5\original} & \scalebox{1.5}{78.9\original} & \scalebox{1.5}{67.6\original} & \scalebox{1.5}{79.6\original} & \scalebox{1.5}{68.1\original} & \scalebox{1.5}{\textbf{82.1}\reproduce} & \scalebox{1.5}{{81.3}\reproduce} \\

& \scalebox{1.5}{RealWorldQA}        & \scalebox{1.5}{-----} & \scalebox{1.5}{68.8\original} & \scalebox{1.5}{60.5\original} & \scalebox{1.5}{64.3\reproduce} & \scalebox{1.5}{69.3\original} & \scalebox{1.5}{68.1\original} & \scalebox{1.5}{62.3\reproduce} & \scalebox{1.5}{70.1\reproduce} & \scalebox{1.5}{\textbf{72.6}\reproduce} \\

& \scalebox{1.5}{MMVP\cite{mmvp}}               & \scalebox{1.5}{CVPR 2024} & \scalebox{1.5}{\textbf{47.3}\original} & \scalebox{1.5}{43.0\original} & \scalebox{1.5}{44.0\original} & \scalebox{1.5}{42.0\original} & \scalebox{1.5}{40.0\original} & \scalebox{1.5}{46.5\original} & \scalebox{1.5}{46.7\reproduce} & \scalebox{1.5}{47.1\reproduce} \\

& \scalebox{1.5}{MMEval-Pro\cite{mmevalpro}}         & \scalebox{1.5}{NAACL 2025} & \scalebox{1.5}{70.6\original} & \scalebox{1.5}{69.4\original} & \scalebox{1.5}{72.2\reproduce} & \scalebox{1.5}{73.2\reproduce} & \scalebox{1.5}{73\original} & \scalebox{1.5}{71.5\reproduce} & \scalebox{1.5}{72.0\reproduce} & \scalebox{1.5}{\textbf{73.6}\reproduce} \\

& \scalebox{1.5}{VMCBench\cite{vmcbench}}           & \scalebox{1.5}{NeurIPS 2025} & \scalebox{1.5}{79.7\original} & \scalebox{1.5}{65.2\original} & \scalebox{1.5}{80.3\original} & \scalebox{1.5}{73.9\fromglm} & \scalebox{1.5}{73.2\original} & \scalebox{1.5}{80.3\original} & \scalebox{1.5}{81.4\reproduce} & \scalebox{1.5}{\textbf{81.5}\reproduce} \\

& \scalebox{1.5}{MMVet\cite{mmvet}}              & \scalebox{1.5}{ICML 2024} & \scalebox{1.5}{66.0\original} & \scalebox{1.5}{72.1\original} & \scalebox{1.5}{65.6\original} & \scalebox{1.5}{\textbf{74.6}\original} & \scalebox{1.5}{73.2\original} & \scalebox{1.5}{66.9\original} & \scalebox{1.5}{67.8\reproduce} & \scalebox{1.5}{74.1\reproduce} \\

& \scalebox{1.5}{MMStar\cite{mmstar}}             & \scalebox{1.5}{NeurIPS 2024} & \scalebox{1.5}{61.7\original} & \scalebox{1.5}{56.2\original} & \scalebox{1.5}{56\reproduce} & \scalebox{1.5}{63\reproduce} & \scalebox{1.5}{65.3\reproduce} & \scalebox{1.5}{60.7\reproduce} & \scalebox{1.5}{65\reproduce} & \scalebox{1.5}{\textbf{66.8}\reproduce} \\

& \scalebox{1.5}{ZeroBench\cite{zerobench}}          & \scalebox{1.5}{Arxiv 2025} & \scalebox{1.5}{13.7\original} & \scalebox{1.5}{14.1\original} & \scalebox{1.5}{16.7\reproduce} & \scalebox{1.5}{15.3\reproduce} & \scalebox{1.5}{15.2\reproduce} & \scalebox{1.5}{17.8\reproduce} & \scalebox{1.5}{18.9\reproduce} & \scalebox{1.5}{\textbf{18.9}\reproduce} \\

\mycmidrulecolumntwo
% Overall 行
% 【核心修改】对第2列使用 multicolumn
& \multicolumn{1}{>{\fontsize{8pt}{10pt}\selectfont}l}{\scalebox{1.5}{Overall}} & \multicolumn{1}{c}{\scalebox{1.5}{}} & \scalebox{1.5}{59.7\original} & \scalebox{1.5}{55.9\original} & \scalebox{1.5}{59.8\original} & \scalebox{1.5}{59.9\original} & \scalebox{1.5}{61.0\original} & \scalebox{1.5}{59.3\original} & \scalebox{1.5}{63.0\original} & \scalebox{1.5}{\textbf{64.5}\original} \\

\midrule
% Hallucination 部分
\multirow{7}{*}{\scalebox{1.5}{\shortstack{Hallucination\\Benchmark}}}

& \scalebox{1.5}{MMHalu\cite{mmhalu}}             & \scalebox{1.5}{ACL 2024} & \scalebox{1.5}{3.50\original} & \scalebox{1.5}{3.26\original} & \scalebox{1.5}{3.57\original} & \scalebox{1.5}{4.06\original} & \scalebox{1.5}{3.95\fromoc} & \scalebox{1.5}{3.00\original} & \scalebox{1.5}{3.30\original} & \scalebox{1.5}{\textbf{4.12}\original} \\

& \scalebox{1.5}{Bingo\cite{bingo}}              & \scalebox{1.5}{NeurIPS 2023} & \scalebox{1.5}{3.65\original} & \scalebox{1.5}{3.65\original} & \scalebox{1.5}{3.58\original} & \scalebox{1.5}{3.67\original} & \scalebox{1.5}{3.59\fromoc} & \scalebox{1.5}{3.45\original} & \scalebox{1.5}{3.3\reproduce} & \scalebox{1.5}{\textbf{3.69}\reproduce} \\

& \scalebox{1.5}{POPE-R\cite{pope}}             & \scalebox{1.5}{EMNLP 2023} & \scalebox{1.5}{82.1\original} & \scalebox{1.5}{84.6\original} & \scalebox{1.5}{88\original} & \scalebox{1.5}{85.5\fromglm} & \scalebox{1.5}{85.0\fromoc} & \scalebox{1.5}{82.4\original} & \scalebox{1.5}{86.9\original} & \scalebox{1.5}{\textbf{88.6}\reproduce} \\

& \scalebox{1.5}{POPE-P\cite{pope}}             & \scalebox{1.5}{EMNLP 2023} & \scalebox{1.5}{83.2\original} & \scalebox{1.5}{84.0\original} & \scalebox{1.5}{85.2\original} & \scalebox{1.5}{81.8\fromglm} & \scalebox{1.5}{85\fromoc} & \scalebox{1.5}{82.5\original} & \scalebox{1.5}{86.7\reproduce} & \scalebox{1.5}{\textbf{87.3}\reproduce} \\

& \scalebox{1.5}{POPE-A\cite{pope}}             & \scalebox{1.5}{EMNLP 2023} & \scalebox{1.5}{83.1\original} & \scalebox{1.5}{82.5\original} & \scalebox{1.5}{84.0\original} & \scalebox{1.5}{82.8\fromglm} & \scalebox{1.5}{84.1\fromoc} & \scalebox{1.5}{79.1\original} & \scalebox{1.5}{87.1\original} & \scalebox{1.5}{\textbf{87.6}\reproduce} \\

& \scalebox{1.5}{HallusionBench\cite{hallusionbench}}     & \scalebox{1.5}{CVPR 2024} & \scalebox{1.5}{60.3\original} & \scalebox{1.5}{60.9\original} & \scalebox{1.5}{59.8\original} & \scalebox{1.5}{56.1\fromglm} & \scalebox{1.5}{56.2\fromoc} & \scalebox{1.5}{63.7\original} & \scalebox{1.5}{\textbf{63.9}\reproduce} & \scalebox{1.5}{62.3\reproduce} \\

\mycmidrulecolumntwo
% Overall 行
& \multicolumn{1}{>{\fontsize{8pt}{10pt}\selectfont}l}{\scalebox{1.5}{Overall}} & \multicolumn{1}{c}{\scalebox{1.5}{}} & \scalebox{1.5}{52.6\original} & \scalebox{1.5}{53.2\original} & \scalebox{1.5}{54\original} & \scalebox{1.5}{52.3\original} & \scalebox{1.5}{53\original} & \scalebox{1.5}{52.4\original} & \scalebox{1.5}{55.2\original} & \scalebox{1.5}{\textbf{55.6}\original} \\

\midrule
% MATH 部分
\multirow{8}{*}{\scalebox{1.5}{\shortstack{Mathematical\\Reasoning}}}

& \scalebox{1.5}{MathVerse\cite{mathverse}}  & \scalebox{1.5}{ECCV 2024} & \scalebox{1.5}{44.5\original} & \scalebox{1.5}{46.4\original} & \scalebox{1.5}{52.4\reproduce} & \scalebox{1.5}{54.2\reproduce} & \scalebox{1.5}{53.3\reproduce} & \scalebox{1.5}{50.3\reproduce} & \scalebox{1.5}{52.1\reproduce} & \scalebox{1.5}{\textbf{54.6}\reproduce} \\

& \scalebox{1.5}{MathVision\cite{mathvision}}   & \scalebox{1.5}{NeurIPS 2024} & \scalebox{1.5}{22.1\original} & \scalebox{1.5}{29.9\original} & \scalebox{1.5}{27.2\reproduce} & \scalebox{1.5}{32.3\reproduce} & \scalebox{1.5}{30.7\reproduce} & \scalebox{1.5}{25.9\reproduce} & \scalebox{1.5}{32.9\reproduce} & \scalebox{1.5}{\textbf{33.7}\reproduce} \\

& \scalebox{1.5}{MathVista\cite{mathvista}}    & \scalebox{1.5}{ICLR 2024} & \scalebox{1.5}{67.8\original} & \scalebox{1.5}{64.1\original} & \scalebox{1.5}{73.5\reproduce} & \scalebox{1.5}{74.9\reproduce} & \scalebox{1.5}{74.8\reproduce} & \scalebox{1.5}{72.3\reproduce} & \scalebox{1.5}{\textbf{75.1}\original} & \scalebox{1.5}{74.9\original} \\

& \scalebox{1.5}{MM-Math\cite{mmmath}}      & \scalebox{1.5}{EMNLP 2024} & \scalebox{1.5}{33.7\original} & \scalebox{1.5}{32.1\original} & \scalebox{1.5}{40.2\reproduce} & \scalebox{1.5}{35.3\reproduce} & \scalebox{1.5}{35.7\reproduce} & \scalebox{1.5}{38.9\reproduce} & \scalebox{1.5}{38.4\original} & \scalebox{1.5}{\textbf{41.1}\original} \\

& \scalebox{1.5}{Geometry3K\cite{geometry3k}} & \scalebox{1.5}{ACL 2021} & \scalebox{1.5}{59.2\original} & \scalebox{1.5}{57.9\original} & \scalebox{1.5}{67\reproduce} & \scalebox{1.5}{67.7\reproduce} & \scalebox{1.5}{68.7\reproduce} & \scalebox{1.5}{60.6\reproduce} & \scalebox{1.5}{68.9\reproduce} & \scalebox{1.5}{\textbf{69.4}\reproduce} \\

& \scalebox{1.5}{VisuLogic\cite{visulogic3k}}    & \scalebox{1.5}{ICLR 2026} & \scalebox{1.5}{26.9\original} & \scalebox{1.5}{24.9\original} & \scalebox{1.5}{26.4\reproduce} & \scalebox{1.5}{27.3\reproduce} & \scalebox{1.5}{28.2\reproduce} & \scalebox{1.5}{29.1\reproduce} & \scalebox{1.5}{28.9\reproduce} & \scalebox{1.5}{\textbf{30.6}\reproduce} \\

& \scalebox{1.5}{MMK12-Math\cite{mmk12}}   & \scalebox{1.5}{Arxiv 2025} & \scalebox{1.5}{58.4\original} & \scalebox{1.5}{44.8\original} & \scalebox{1.5}{52.1\reproduce} & \scalebox{1.5}{51.3\reproduce} & \scalebox{1.5}{63.7\reproduce} & \scalebox{1.5}{63\reproduce} & \scalebox{1.5}{63.9\reproduce} & \scalebox{1.5}{\textbf{64.7}\reproduce} \\

\mycmidrulecolumntwo
% Overall 行
& \multicolumn{1}{>{\fontsize{8pt}{10pt}\selectfont}l}{\scalebox{1.5}{Overall}} & \multicolumn{1}{c}{\scalebox{1.5}{}} & \scalebox{1.5}{44.7\original} & \scalebox{1.5}{42.9\original} & \scalebox{1.5}{48.4\original} & \scalebox{1.5}{49.0\original} & \scalebox{1.5}{50.7\original} & \scalebox{1.5}{48.6\original} & \scalebox{1.5}{51.5\original} & \scalebox{1.5}{\textbf{52.6}\original} \\

\midrule
% Scientific 部分
\multirow{6}{*}{\scalebox{1.5}{\shortstack{Scientific\\Reasoning}}}

& \scalebox{1.5}{MMK12-Phys\cite{mmk12}}     & \scalebox{1.5}{Arxiv 2025} & \scalebox{1.5}{45.4\original} & \scalebox{1.5}{33.8\original} & \scalebox{1.5}{47.3\reproduce} & \scalebox{1.5}{47.2\reproduce} & \scalebox{1.5}{43.2\reproduce} & \scalebox{1.5}{53.8\reproduce} & \scalebox{1.5}{46.9\reproduce} & \scalebox{1.5}{\textbf{56.9}\reproduce} \\

& \scalebox{1.5}{MMK12-Chem\cite{mmk12}}    & \scalebox{1.5}{Arxiv 2025} & \scalebox{1.5}{56.4\original} & \scalebox{1.5}{39.8\original} & \scalebox{1.5}{55.4\original} & \scalebox{1.5}{57.4\original} & \scalebox{1.5}{57.5\original} & \scalebox{1.5}{60.6\original} & \scalebox{1.5}{57.9\original} & \scalebox{1.5}{\textbf{63.7}\original} \\

& \scalebox{1.5}{MMK12-Bio\cite{mmk12}} & \scalebox{1.5}{Arxiv 2025} & \scalebox{1.5}{54\original} & \scalebox{1.5}{40.8\original} & \scalebox{1.5}{57.9\original} & \scalebox{1.5}{64.8\original} & \scalebox{1.5}{61.3\original} & \scalebox{1.5}{65\original} & \scalebox{1.5}{62.9\original} & \scalebox{1.5}{\textbf{65.7}\original} \\

& \scalebox{1.5}{PhyUniBench\cite{phyunibench}}     & \scalebox{1.5}{Arxiv 2025} & \scalebox{1.5}{17.3\original} & \scalebox{1.5}{19.6\original} & \scalebox{1.5}{18.9\original} & \scalebox{1.5}{19.2\original} & \scalebox{1.5}{18.1\original} & \scalebox{1.5}{18.3\original} & \scalebox{1.5}{18.1\original} & \scalebox{1.5}{\textbf{19.9}\original} \\

& \scalebox{1.5}{PhyX\cite{phyx}}     & \scalebox{1.5}{Arxiv 2025} & \scalebox{1.5}{30.6\original} & \scalebox{1.5}{32.1\original} & \scalebox{1.5}{41.5\reproduce} & \scalebox{1.5}{37.4\reproduce} & \scalebox{1.5}{41.5\reproduce} & \scalebox{1.5}{39.5\reproduce} & \scalebox{1.5}{39.7\reproduce} & \scalebox{1.5}{\textbf{42.9}\reproduce} \\

\mycmidrulecolumntwo
% Overall 行
& \multicolumn{1}{>{\fontsize{8pt}{10pt}\selectfont}l}{\scalebox{1.5}{Overall}} & \multicolumn{1}{c}{\scalebox{1.5}{}} & \scalebox{1.5}{40.7\original} & \scalebox{1.5}{33.2\original} & \scalebox{1.5}{44.2\original} & \scalebox{1.5}{45.2\original} & \scalebox{1.5}{44.3\original} & \scalebox{1.5}{47.4\original} & \scalebox{1.5}{45.1\original} & \scalebox{1.5}{\textbf{49.8}\original} \\

\bottomrule
\end{tabular}
}
\vspace{2mm}
\end{threeparttable}
\vspace{-4mm}
\end{table*}

\textbf{3. Reflection Format Reward ($R_{reflect}$):} 
We emphasize that $R_{reflect}$ is a lightweight \emph{format} reward, rather than a correctness improvement bonus. 
Its role is to encourage a natural reflection marker (e.g., ``Wait...'') to appear when FRM is triggered, so that the model internalizes a coherent pause and check habit instead of producing command following boilerplate.
Let $\mathbb{I}_{\mathrm{FRM}}(Y)$ indicate whether FRM is triggered for the rollout, and let $n_{\tau_{nat}}(Y)$ count the occurrences of the natural marker $\tau_{nat}$ in $Y$.
Let $t_r$ denote the position of the first occurrence of $\tau_{nat}$ in $Y$.
If $n_{\tau_{nat}}(Y)=0$, we set $\bar S^{mid}_{refl}(Y)=0$.
We measure visual reengagement after reflection by
\begin{equation}
\bar S^{mid}_{refl}(Y)=\frac{1}{K}\sum_{t=t_r}^{t_r+K-1}\bar S^{mid}_t,
\label{eq24}
\end{equation}
where $K$ is a fixed constant.
We define
\begin{equation}
\begin{aligned}
R_{reflect}
&= 0.4\,\mathbb{I}_{\mathrm{FRM}}(Y)\,
\mathbb{I}\!\left[1 \le n_{\tau_{nat}}(Y) \le 3\right] \\
&\quad \mathbb{I}\!\left[\bar S^{mid}_{refl}(Y)\ge \tau_{mid}\right].
\end{aligned}
\label{eq25}
\end{equation}
This reward encourages a natural reflection cue while avoiding degenerate overuse.

\textbf{Hybrid Loss Function.} 
The final objective combines GRPO\cite{paper6} with the auxiliary SFT loss over the replay buffer (Sec.~\ref{4.4}):
\begin{equation}
\mathcal{J}(\theta) = \mathcal{L}_{GRPO}(\theta) + \gamma \mathcal{L}_{\mathrm{SFT}}(\theta).
\label{eq26}
\end{equation}
By jointly optimizing this hybrid objective, V-STAR translates the intrinsic capabilities of visual backtracking and logical reflection into parameter updates.

\section{Experiments}
 \label{5}

%\subsection{Experimental Setup}
%\label{subsec:Experimental Setup}

\subsection{Baselines and Model Implementation}
\label{5.1}
We build V-STAR based on the Qwen2.5-VL-7B\cite{qwen25vl} architecture. To evaluate its performance, we compare it against several representative Multimodal Large Reasoning Models (MLRMs). These include general purpose models like R1-Onevision-7B\cite{r1onevision} and Vision-R1-7B\cite{visionr1}, as well as models specialized for reasoning tasks, such as VL-Rethinker-7B\cite{vlrethinker}, VL-Cogito-7B\cite{vlcogito}, OpenVLThinker-7B\cite{openvlthinker}, and ThinkLite-VL-7B\cite{thinklitevl}.

\subsection{Evaluation Protocols and Benchmarks}
\label{5.2}
To ensure a holistic assessment of V-STAR’s capabilities, we follow the official evaluation protocol of each benchmark and evaluate on a suite of benchmarks spanning four domains: (i) General Reasoning and Understanding, (ii) Hallucination Benchmarks, (iii) Mathematical Reasoning, and (iv) Scientific Reasoning. The first two domains test perceptual robustness and factual faithfulness in open-ended vision-language scenarios, with an emphasis on fine-grained perception and object-level hallucination diagnosis. The latter two domains focus on structured STEM reasoning to verify that V-STAR preserves rigorous, verifiable logic and answer correctness while improving visual grounding. Table~\ref{tab1} summarizes the overall performance comparison across all benchmarks, and detailed benchmark profiles, including visual input formats, metrics, and evaluation focus, are provided in Appendix Table 1.

As shown in Table~\ref{tab1}, V-STAR exhibits a clear advantage across domains rather than a narrow gain on a single task family. Its improvements are especially consistent on hallucination and structured reasoning benchmarks, where visual grounding is crucial for maintaining faithful long-chain reasoning. Meanwhile, the results are not uniformly saturated on every benchmark, suggesting that V-STAR does not simply overfit to one evaluation style, but instead improves robustness across diverse visual reasoning settings.

\subsection{Ablation Study}
\label{5.3}

\textbf{Effect of Key Components.} We analyze the contribution of each module in V-STAR as reported in Table~\ref{tab2}. A clear pattern emerges from the ablation results. The base model with simple outcome supervision brings only limited gains over Qwen2.5-VL-7B\cite{qwen25vl}, suggesting that reward on final answers alone is insufficient to correct the internal grounding failure. In contrast, HVAR yields a larger improvement on hallucination-sensitive benchmarks such as MMHalu\cite{mmhalu} and also improves RealWorld, indicating that explicit visual reinforcement is particularly effective when the model must stay anchored to image evidence during open-ended reasoning. FRM shows a similarly strong effect on structure-intensive tasks such as Geometry3K\cite{geometry3k}, where interrupting erroneous reasoning momentum is especially important. When the two components are combined, the improvements become consistently stronger across all benchmarks, which suggests that visual re-anchoring and reflection-based trajectory correction are complementary rather than redundant. Overall, the ablation results indicate that V-STAR benefits not from a single trick, but from the interaction between grounded visual recovery and trajectory-level self-correction.

 \begin{table}[t]
\small
 \centering
 \setlength{\tabcolsep}{2pt}
 \renewcommand\arraystretch{1.15}
 \caption{\textbf{Component Ablation Study.}We analyze the impact of adding the Visual Reward (HVAR) and Forced Reflection (FRM) to the base model. Results show that both components are essential, and combining them yields the highest improvement across all metrics.}
 \scalebox{0.96}{
 \begin{tabular}{l||cccc}
 \hline\hline
 \rowcolor{lightgray}
  & \multicolumn{4}{c}{Benchmarks} \\
 \cline{2-5}
 \rowcolor{lightgray}
 Methods & MMEval-Pro & MMHalu & Geo3K & RealWorld \\
 \hline\hline
 Qwen2.5VL-7B     & 70.6 & 3.50 & 59.2 & 68.8 \\
 Reward on Answer & 71.3 & 3.81 & 62.4 & 69.7 \\
 RL-HVAR-only     & 72.9 & 3.92 & 67.3 & 71.7 \\
 RL-FRM-only      & 72.8 & 3.99 & 67.0 & 71.6 \\
 V-STAR           & \textbf{73.6} & \textbf{4.12} & \textbf{68.9} & \textbf{72.6} \\
 \hline\hline
 \end{tabular}}
 \label{tab2}
 \vspace{-4mm}
 \end{table}

%\begin{table*}[htbp]
%\small
%\centering
%\setlength{\tabcolsep}{12pt} 
%\renewcommand\arraystretch{1.2}
%\caption{Ablation study of different components on benchmarks.}
%\begin{tabular}{l||cccc}
%\hline\hline
%\rowcolor{lightgray}
% & \multicolumn{4}{c}{Benchmarks} \\
%\cline{2-5}
%\rowcolor{lightgray}
%Methods & MMEval-Pro & MMHalu & Geometry3K & RealWorldQA \\
%\hline\hline
%Qwen2.5VL-7B     & 70.6 & 3.50 & 59.2 & 68.8 \\
%Reward on Answer & 71.3 & 3.81 & 62.4 & 69.7 \\
%RL-HVAR-only     & 72.9 & 3.92 & 67.3 & 71.7 \\
%RL-FRM-only      & 72.8 & 3.99 & 67.0 & 71.6 \\
%V-STAR           & \textbf{73.6} & \textbf{4.12} & \textbf{68.9} & %\textbf{72.6} \\
%\hline\hline
%\end{tabular}
%\label{tab:ablation_components}
%\end{table*}

\begin{table}[!t]
\small
\centering
\setlength{\tabcolsep}{2.0pt}
\renewcommand\arraystretch{1.2}
\caption{\textbf{Ablation on Data Selection Strategy.} We compare different data selection ratios ($\eta$) and strategies. At matched data scales, the \emph{Refine} strategy, which applies curated correctness-preserving selection, consistently outperforms random selection across all evaluated benchmarks.}
\resizebox{\columnwidth}{!}{
\begin{tabular}{c|l|c||cccc}
\hline\hline
\rowcolor{lightgray}
 & & & \multicolumn{4}{c}{Benchmarks} \\
\cline{4-7}
\rowcolor{lightgray}
\multirow{-2}{*}{$\eta$} & \multirow{-2}{*}{Method} & \multirow{-2}{*}{Data size}
 & MMEval-Pro & MMHalu & Geo3K & RealWorld \\
\hline\hline
0 & Base & - & 70.6 & 3.50 & 59.2 & 68.8 \\
\hline
\multirow{2}{*}{0.2} & Refine & 8k & 71.6 & 3.69 & 63.1 & 70.2 \\
 & Random & 8k & 71.0 & 3.58 & 61.4 & 69.4 \\
\hline
\multirow{2}{*}{0.4} & Refine & 16k & 72.3 & 3.81 & 65.9 & 71.1 \\
 & Random & 16k & 71.6 & 3.74 & 63.6 & 70.3 \\
\hline
\multirow{2}{*}{0.6} & Refine & 24k & 72.6 & 3.99 & 67.4 & 71.9 \\
 & Random & 24k & 72.1 & 3.92 & 65.3 & 71.0 \\
\hline\hline
\end{tabular}}
\label{tab3}
\vspace{-4mm}
\end{table}

\begin{figure}[t]
    \centering
    \vskip -0.00in
    \hspace*{-1.6mm}\includegraphics[width=1.0\linewidth]{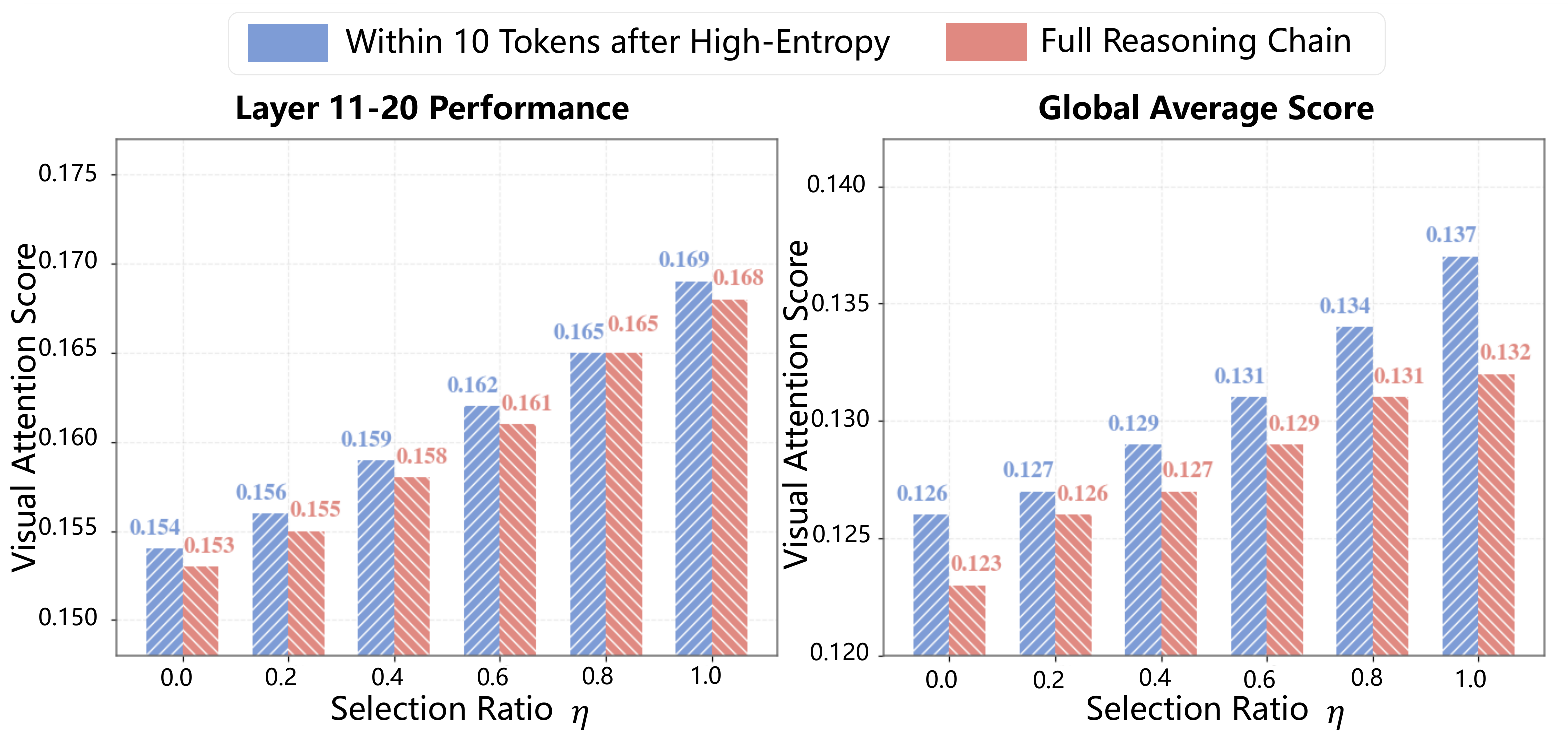}
    \vskip -0.15in
    \caption{\textbf{Selection ratio $\eta$ in data curation.}
    Visual attention scores are reported for the anchoring layers (11--20) and the full network average, measured both in a pivot local window (10 tokens after high entropy pivots) and over the full chain.
    With increasing $\eta$, the visual attention scores steadily rise under both measurement windows, including the pivot local window and the full reasoning chain.}
    \label{fig10}
\vskip -0.0in
\end{figure}

\textbf{Effect of Data Selection Strategy.} We test how dataset scale and selection strategy affect performance. As shown in Table~\ref{tab3}, performance improves steadily as the training set grows from 8k to 24k samples, and the gains gradually diminish at larger scales, suggesting a trend toward saturation rather than indefinite benefits from adding more data. More importantly, at each matched data scale, Refine consistently outperforms Random across all benchmarks. This suggests that the advantage cannot be explained by sample quantity alone, but also by the higher quality supervision induced by curated selection that preserves correctness. In other words, training on curated final-correct reasoning trajectories provides a more effective learning signal than using unfiltered reasoning traces. These results highlight that supervision quality matters in addition to raw data volume. The trend further suggests that carefully curated trajectories with correct final answers provide useful signals for learning more stable reasoning behavior and may help the model better handle unstable intermediate reasoning states.

\textbf{Visual Attention Recovery.} To validate the mechanism of HVAR, we analyze the visual attention shifts in intermediate layers from layer 11 to layer 20. Fig.~\ref{fig10} reveals two notable phenomena. First, visual attention increases monotonically as the selection ratio $\eta$ grows, indicating that better curated data steadily strengthens the model’s tendency to consult visual evidence. Second, this increase is more pronounced in the local window around pivots than in the full-chain average, showing that the recovered attention is not uniformly distributed but concentrated around high-uncertainty reasoning steps. This observation is important because it suggests that HVAR does not merely raise global visual activity, but preferentially restores attention exactly at the points where hallucinations are most likely to emerge. The result is consistent with our motivation that effective visual grounding should be strongest at cognitively unstable steps rather than uniformly spread across the entire reasoning chain.

\begin{figure}[t]
    \centering
    \vskip -0.00in
    \includegraphics[width=1.03\linewidth]{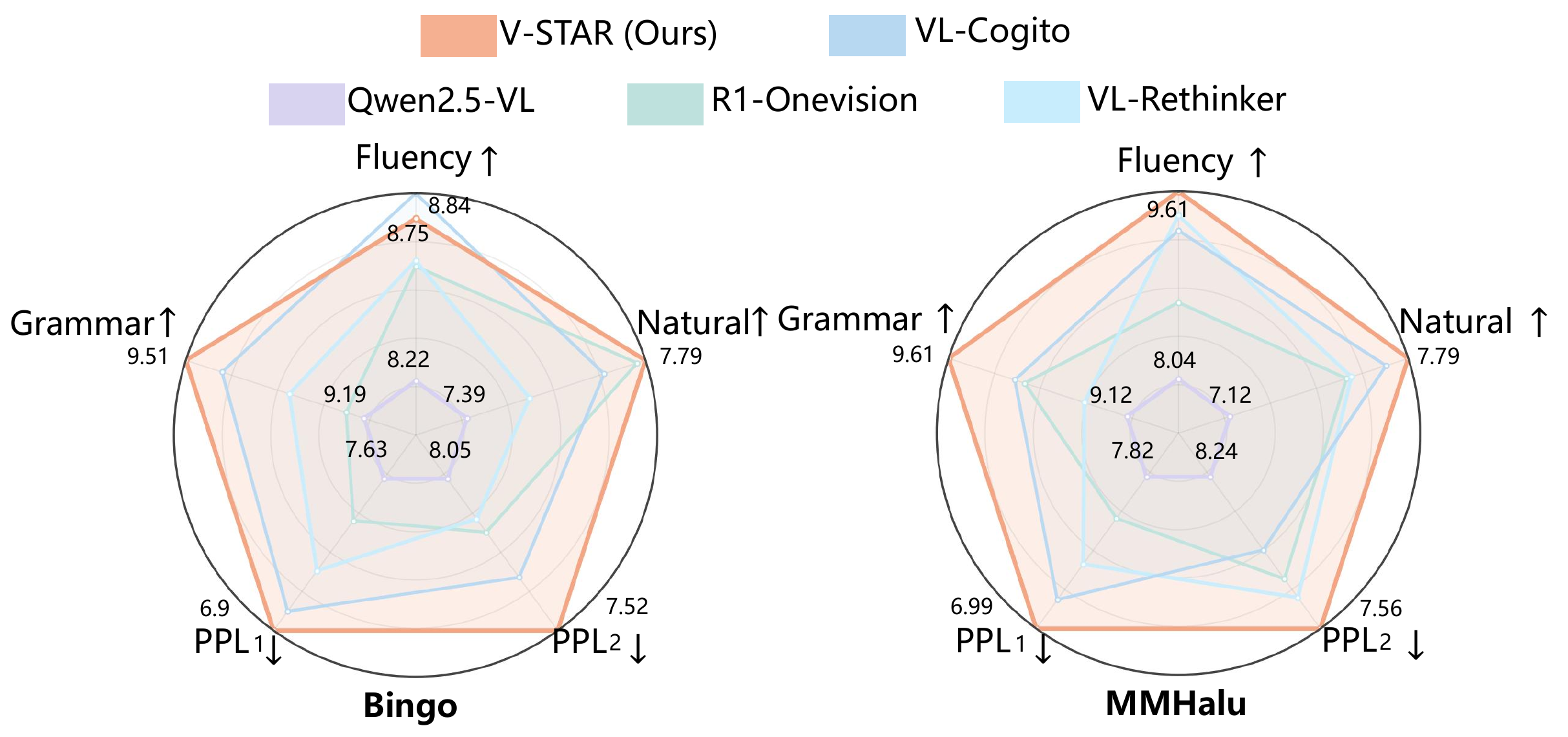}
    \vskip -0.1in
    \caption{\textbf{Linguistic quality of reflection outputs.}
    We report automatic text quality scores (Naturalness, Fluency, Grammar; $\uparrow$) and perplexities (PPL1, PPL2; $\downarrow$) on Bingo\cite{bingo} and MMHalu\cite{mmhalu}.
    PPL1 and PPL2 are calculated using GPT-2, while the ratings for Grammar, Fluency, and Naturalness are provided by GPT-5.
    While enabling explicit reflection, V-STAR shows no degradation in language quality and achieves better results on multiple metrics.}
    \label{fig12}
\vskip -0.3in
\end{figure}

\textbf{Linguistic Quality of Reflection.} A critical concern in RL-based generation is the potential degradation of linguistic fluency. We therefore evaluate whether the model still produces natural text using GPT-based scores including Naturalness, Fluency, and Grammar, together with Perplexity metrics. As shown in Fig.~\ref{fig12}, V-STAR does not exhibit the common trade-off between stronger control and worse language quality. Instead, it remains competitive or better on fluency-related scores while also achieving lower perplexity on both benchmarks. This means that the introduced reflection behavior is not a brittle decoding artifact or a rigid template memorization pattern; rather, it is integrated into generation in a linguistically stable manner.

This finding is important for understanding the nature of the gains brought by V-STAR. If explicit reflection merely forced the model to insert repetitive self-check phrases, one would expect a deterioration in naturalness, fluency, or grammatical consistency. However, the results suggest the opposite: reflection is absorbed into the model’s generation behavior without compromising readability. In this sense, V-STAR does not simply superimpose an external control pattern on top of an unchanged language model. Instead, it appears to reshape the overall reasoning style into one that is both more grounded and more coherent. This makes the observed performance improvements more convincing, since they cannot be explained away as superficial prompting behavior or decoding driven by templates.

\subsection{Comparisons to State of the Arts}
\label{5.4}

\begin{figure*}[htbp]
\vskip -0.00in
\centering
\centerline{\includegraphics[width=2.05\columnwidth]{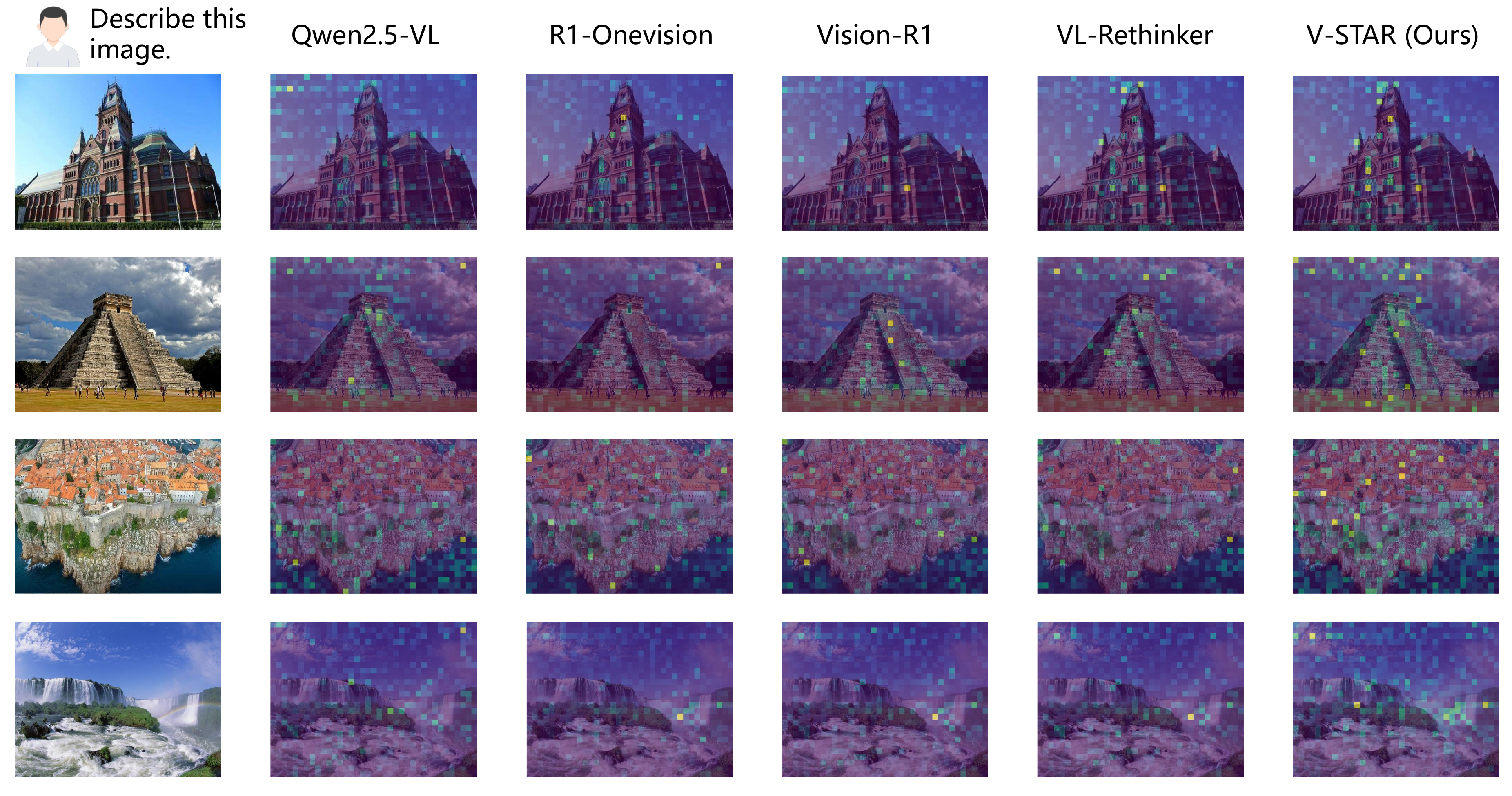}}
\vskip -0.1in
\caption{\textbf{Attention heatmaps under identical prompting.}
Compared with representative baselines, V-STAR allocates more token to image attention to visual evidence during generation and concentrates this attention on semantically relevant regions, while suppressing background dominated activation.
This pattern is consistent with stronger visual anchoring.}
\label{fig11}
\vskip -0.1in
\end{figure*}

\textbf{Qualitative Analysis.} We compare attention heatmaps to visually interpret the impact of V-STAR. As shown in Fig.~\ref{fig11}, the difference is not only in the amount of image attention, but also in its spatial distribution. Baseline models often either diffuse their attention over broad background regions or gradually lose concentration as reasoning proceeds. In contrast, V-STAR allocates denser and more localized attention to semantically relevant regions, especially near logical turning points. This suggests that our method changes visual processing from passive exposure to selective evidence seeking, which is more consistent with grounded reasoning. The qualitative examples therefore support the quantitative attention analysis by showing that V-STAR not only looks more, but also looks more precisely at the regions that matter for the current reasoning step.

\begin{table}[t]
\small
\centering
\setlength{\tabcolsep}{3.5pt}
\renewcommand\arraystretch{1.15}
\caption{\textbf{Robustness on Object Hallucination (POPE).}We test model performance on the POPE benchmark using different decoding strategies (Greedy vs. Sampling). V-STAR maintains high stability and accuracy across all splits (Random, Popular, Adversarial), effectively reducing object hallucinations.}
\scalebox{0.92}{
\begin{tabular}{l|l||cc|cc|cc}
\hline\hline
\rowcolor{lightgray}
 & & \multicolumn{2}{c}{Random} & \multicolumn{2}{c}{Popular} & \multicolumn{2}{c}{Adversarial} \\
\cline{3-8}
\rowcolor{lightgray}
\multirow{-2}{*}{Model} & \multirow{-2}{*}{Decoding}
 & Acc & F1 & Acc & F1 & Acc & F1 \\
\hline\hline
\multirow{2}{*}{Qwen2.5VL-7B}
 & Greedy   & 81.4 & 80.6 & 82.5 & 81.7 & 82.3 & 81.8 \\
 & Sampling & 82.1 & 81.3 & 83.2 & 82.4 & 83.1 & 82.6 \\
\hline
\multirow{2}{*}{R1-Onevision}
 & Greedy   & 83.9 & 83.1 & 83.3 & 82.4 & 81.8 & 81.2 \\
 & Sampling & 84.6 & 83.8 & 84.0 & 83.1 & 82.5 & 81.9 \\
\hline
\multirow{2}{*}{VL-Rethinker}
 & Greedy   & 83.3 & 82.4 & 82.8 & 81.9 & 81.2 & 80.6 \\
 & Sampling & 85.5 & 83.2 & 81.8 & 82.6 & 82.8 & 81.2 \\
\hline
\multirow{2}{*}{VL-Cogito}
 & Greedy   & 84.0 & 84.4 & 83.5 & 84.1 & 84.1 & 83.5 \\
 & Sampling & 85.0 & 85.0 & 85.0 & 84.8 & 81.9 & 84.1 \\
\hline
\multirow{2}{*}{V-STAR}
 & Greedy   & 88.6 & 87.3 & 86.7 & 86.0 & 86.9 & 86.3 \\
 & Sampling & 87.9 & 88.0 & 87.3 & 86.6 & 87.6 & 87.0 \\
\hline\hline
\end{tabular}}
\label{tab4}
\end{table}

\textbf{Specific Analysis on POPE\cite{pope}.} We perform a detailed analysis on the POPE benchmark to check robustness against object hallucination as detailed in Table~\ref{tab4}. Two phenomena can be observed. First, V-STAR achieves the best accuracy across Random, Popular, and Adversarial splits, indicating that its gains are not restricted to a single distribution. Second, this advantage remains stable under both Greedy and Sampling decoding, suggesting that the reduction of hallucination is intrinsic to the model rather than sensitive to decoding. Notably, the margin becomes more evident on the difficult Adversarial split, which indicates that V-STAR is especially effective when the model is exposed to misleading object priors and must rely more heavily on visual evidence. These results show that the benefit of V-STAR is not limited to typical settings, but becomes particularly clear in scenarios where grounding at the object level is most challenging.

\begin{figure}[t]
    \centering
    \vskip -0.00in
    \hspace*{-1.5mm}\includegraphics[width=1.0\linewidth]{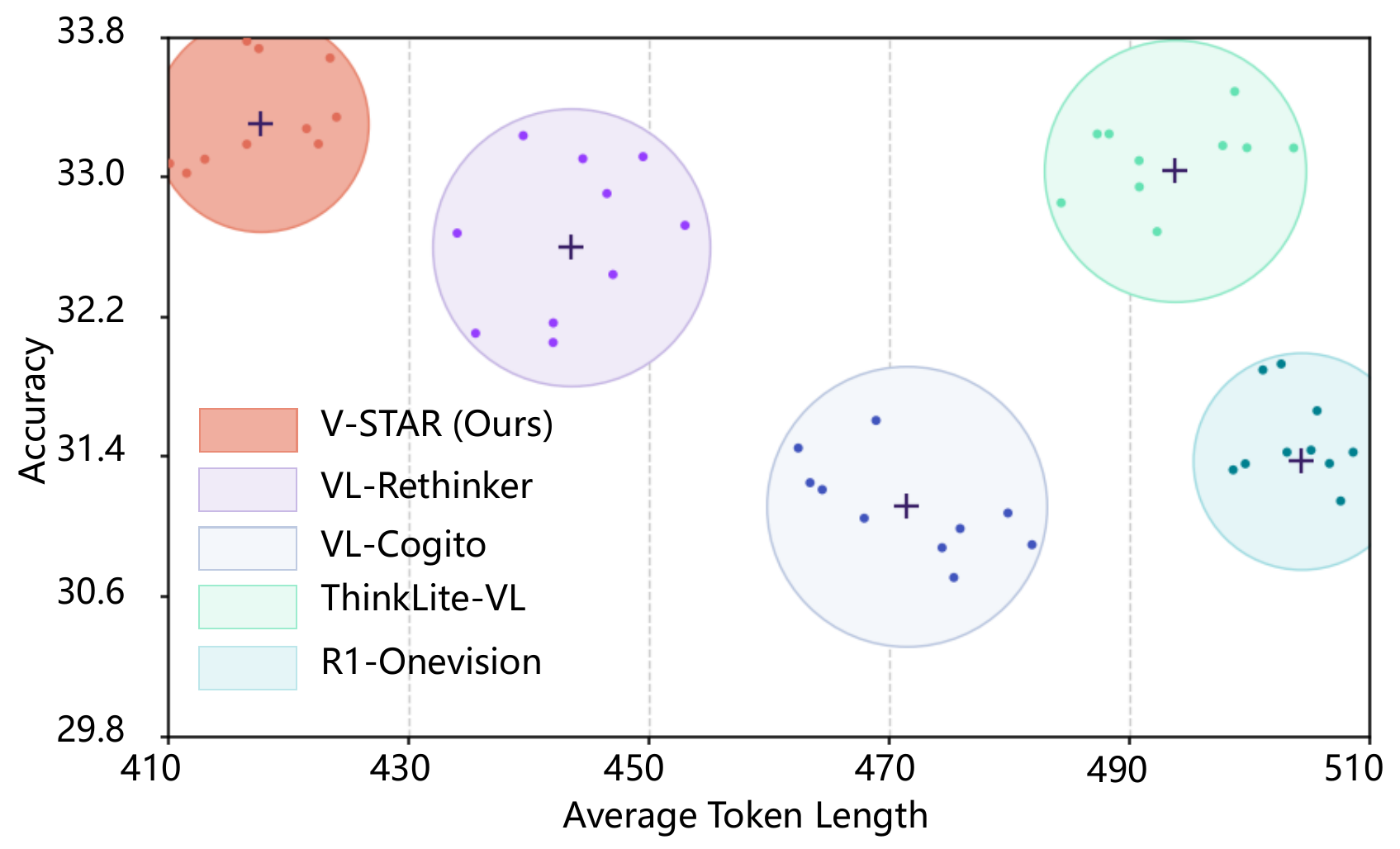}
    \vskip -0.1in
    \caption{\textbf{Accuracy length trade-off on MathVision.}
We plot accuracy versus average generated token length under the same evaluation protocol.
V-STAR achieves higher accuracy with shorter generations, supporting improved reasoning efficiency.}
    \label{fig13}
\vskip -0.0in
\end{figure}

\textbf{Reasoning Efficiency.} We evaluate reasoning efficiency on MathVision\cite{mathvision} by comparing V-STAR with various reasoning models, as shown in Fig.~\ref{fig13}. The result reveals a favorable accuracy-length trade-off: V-STAR lies in the upper left region of the plot, achieving the best accuracy while using fewer generated tokens than competing reasoning models. This suggests that the improvement does not come from simply extending the chain of thought. Instead, stronger visual anchoring appears to reduce unnecessary semantic wandering, allowing the model to arrive at correct solutions with shorter and more targeted reasoning traces. This is an important observation because it shows that better grounding can improve both correctness and efficiency at the same time. It also suggests that V-STAR solves problems more directly by making more effective use of visual evidence during reasoning.

\begin{figure}[t]
    \centering
    \vskip -0.00in
    \hspace*{-3mm}\includegraphics[width=1.00\linewidth]{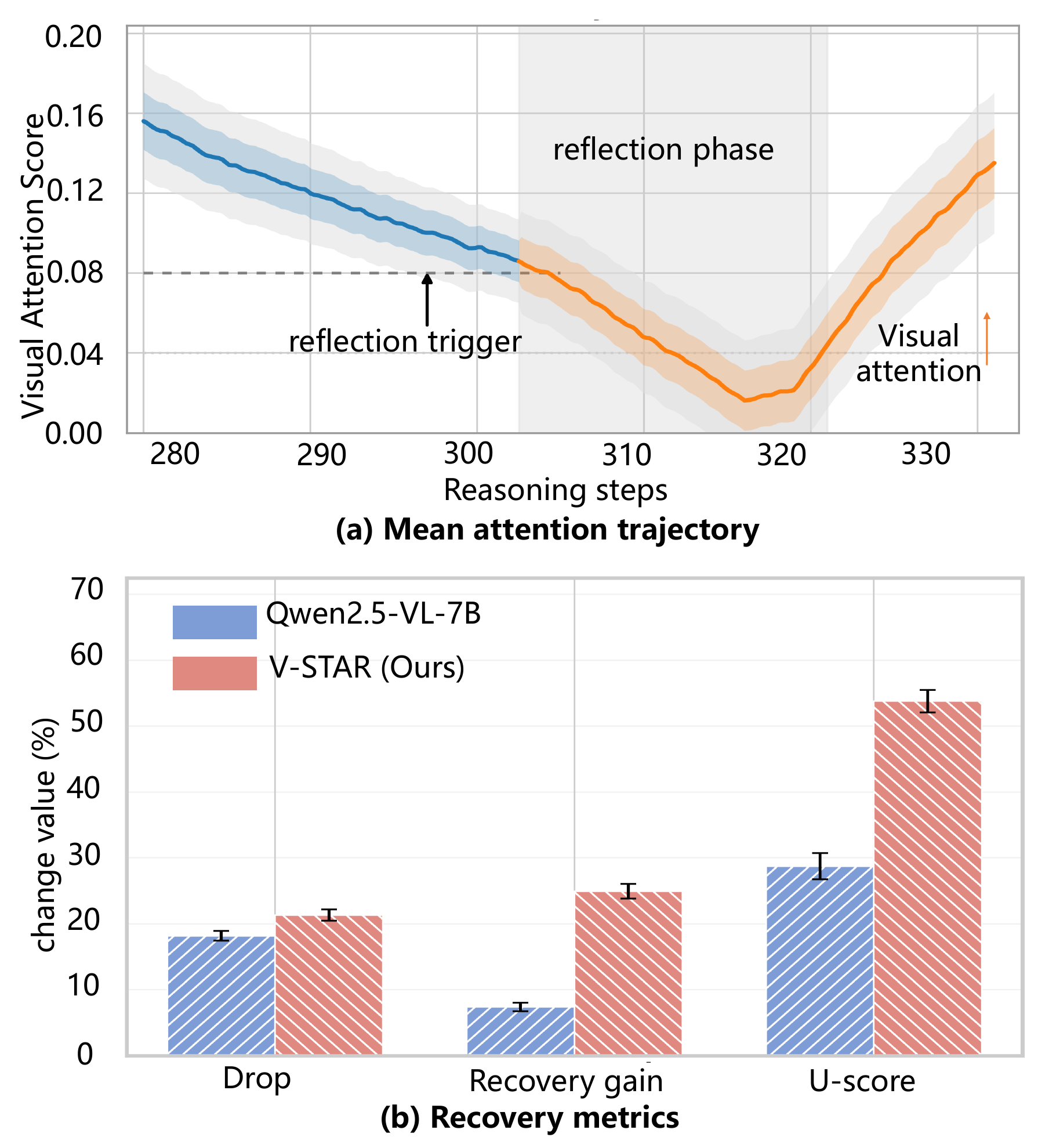}
    \vskip -0.00in
    \caption{\textbf{Visual attention recovery during reflection.}
    \textbf{(a)} Mean visual attention trajectory of V-STAR around the reflection trigger. After entering the reflection phase, the visual attention score shows a clear rebound.
    \textbf{(b)} Recovery metrics, including attention drop, recovery gain, and U score, comparing Qwen2.5-VL-7B with V-STAR. V-STAR achieves higher recovery gain and a higher U score, consistent with more grounded reflection.}
    \label{fig14}
\vskip -0.0in
\end{figure}

\textbf{Verification of Reflection Mechanism.} We trace the visual attention score along the reasoning steps to verify whether the model actually re-checks the image during reflection. As shown in Fig.~\ref{fig14}(a), V-STAR exhibits a clear U-shaped attention trajectory: visual attention first drops before the reflection trigger and then rebounds sharply after entering the reflection phase. This rebound is important because it distinguishes genuine visual re-verification from purely textual self-talk. We further summarize this behavior in Fig.~\ref{fig14}(b) using the U-score:
\begin{equation}
\mathrm{U\text{-}score(\%)} = 100 \cdot \frac{\Delta_{\mathrm{rec}}}{\Delta_{\mathrm{drop}}+\Delta_{\mathrm{rec}}+\epsilon},
\label{eq27}
\end{equation}
where $\Delta_{\mathrm{drop}}$ is the attention decrease and $\Delta_{\mathrm{rec}}$ is the subsequent recovery. V-STAR achieves higher recovery gain and U-score than Qwen2.5-VL-7B, indicating that its reflection stage is accompanied by actual visual re-engagement rather than superficial verbal correction.

\section{Conclusion}
\label{sec:Conclusion}

\newcolumntype{L}[1]{>{\raggedright\arraybackslash}p{#1}}
\newcolumntype{C}[1]{>{\centering\arraybackslash}p{#1}}
\newcolumntype{Y}{>{\raggedright\arraybackslash}X}
\renewcommand\theadfont{\bfseries}

This paper presents a mechanistic account of hallucinations in multimodal large reasoning models by identifying the Reasoning Vision Truth Disconnect, where long chain reasoning increasingly underutilizes visual evidence and drifts toward language priors. We show that errors are tightly coupled with high entropy cognitive bifurcation points, and that more than 70 percent of hallucinations emerge immediately after these unstable pivots. We further localize the failure to intermediate layers, where visual semantic anchoring collapses and visual attention can drop to near zero during high uncertainty transitions. Finally, we uncover a pseudo reflection paradox: even when the model produces reflective cues, visual attention does not rebound, suggesting that verbal self audit can occur without renewed visual grounding.

Guided by these findings, we propose V-STAR, a holistic training paradigm that shifts hallucination mitigation from outcome level supervision toward fine grained internal attention guidance. V-STAR combines a Hierarchical Visual Attention Reward that incentivizes visual re anchoring at high entropy pivots, and a Forced Reflection Mechanism that edits trajectories to disrupt cognitive inertia and internalize self checking behavior. Extensive experiments and ablations show that both components are essential, and that combining them yields the strongest overall improvements. Together, our results suggest that reliable multimodal reasoning depends not only on better final answers, but also on maintaining effective visual grounding dynamics precisely at the moments of greatest uncertainty.

\ifCLASSOPTIONcaptionsoff
  \newpage
\fi


% \newpage
% \newpage

{
\bibliographystyle{IEEEtran}
\bibliography{ref}
}

%%%%%%%%%%%%%%%%%%%%%%%%%%%%%%%%%%%%%%%%%%%%%%%%%%%%%%%%%%%%%%%%%%%%%%%%%%%
%----------------------
%----------------------
%----------------------
% \newpage
% \input{sec/bios}

% \vfill

\end{document}